\PassOptionsToPackage{svgnames,table,table,dvipsnames}{xcolor}

\documentclass[journal]{vgtc}                     % final (journal style)
\onlineid{0}

\vgtccategory{Research}

\vgtcpapertype{evaluation}

\title{Large-Scale Evaluation of Topic Models and Dimensionality Reduction Methods for 2D Text Spatialization}

\shortauthortitle{Atzberger, Cech, \MakeLowercase{\textit{et al.}}: Large-Scale Evaluation of Topic Models and Dimensionality Reductions for 2D Text Spatialization}
\addtocounter{footnote}{1}
\author{%
  \authororcid{Daniel Atzberger\footnotemark[\value{footnote}]}{0000-0002-5409-7843},
  \authororcid{Tim Cech\footnotemark[\value{footnote}]}{0000-0001-8688-2419},
  \authororcid{Matthias Trapp}{0000-0003-3861-5759},
  \authororcid{Rico Richter}{0000-0001-5523-3694},\\
  \authororcid{Willy Scheibel}{0000-0002-7885-9857},
  \authororcid{Jürgen Döllner}{0000-0002-8981-8583}, and
  \authororcid{Tobias Schreck}{0000-0003-0778-8665}
}

\authorfooter{
\item[] $^*$Both authors contributed equally to this work.
\item
    Daniel Atzberger, Matthias Trapp, and Willy Scheibel are with University of Potsdam, Digital Engineering Faculty, Hasso Plattner Institute. E-mail: \{firstname.lastname\}@hpi.uni-potsdam.de
\item
    Tim Cech, Rico Richter, and Jürgen Döllner are with University of Potsdam, Digital Engineering Faculty. E-mail: \{tcech,rico.richter.1,doellner\}@uni-potsdam.de
\item
    Tobias Schreck is with Graz University of Technology. E-mail: tobias.schreck@cgv.tugraz.at
}

\abstract{Topic models are a class of unsupervised learning algorithms for detecting the semantic structure within a text corpus. Together with a subsequent dimensionality reduction algorithm, topic models can be used for deriving spatializations for text corpora as two-dimensional scatter plots, reflecting semantic similarity between the documents and supporting corpus analysis. Although the choice of the topic model, the dimensionality reduction, and their underlying hyperparameters significantly impact the resulting layout, it is unknown which particular combinations result in high-quality layouts with respect to accuracy and perception metrics.
To investigate the effectiveness of topic models and dimensionality reduction methods for the spatialization of corpora as two-dimensional scatter plots (or basis for landscape-type visualizations), we present a large-scale, benchmark-based computational evaluation. Our evaluation consists of (1) a set of corpora, (2) a set of layout algorithms that are combinations of topic models and dimensionality reductions, and (3) quality metrics for quantifying the resulting layout. The corpora are given as  document-term matrices, and each document is assigned to a thematic class.
The chosen metrics quantify the preservation of local and global properties and the perceptual effectiveness of the two-dimensional scatter plots. By evaluating the benchmark on a computing cluster, we derived a multivariate dataset with over \numprint{45000} individual layouts and corresponding quality metrics. Based on the results, we propose guidelines for the effective design of text spatializations that are based on topic models and dimensionality reductions.
As a main result, we show that interpretable topic models are beneficial for capturing the structure of text corpora. We furthermore recommend the use of t-SNE as a subsequent dimensionality reduction.
}

\teaser{
    \centering
\newcommand{\layoutalphalabel}[2]{
    \path[myroundness,myborder] ($(#1.south west)+(0pt,-0.7*\labelsize\baselineskip)+(0pt,\labelsize\baselineskip)$) -- ++(+\labellength\baselineskip,0pt) -- ++(0pt,-\labelsize\baselineskip);
    \node[anchor=center,mylabel] (#1-alpha) at ($(#1.south west)+(0pt,-0.7*\labelsize\baselineskip)+(0.5*\labellength\baselineskip,0.5*\labelsize\baselineskip)$) { $\alpha \approx$ \numprint{#2} };
}
\newcommand{\layoutbetalabel}[2]{
    \path[myroundness,myborder] ($(#1.south east)+(0pt,-0.7*\labelsize\baselineskip)+(0pt,\labelsize\baselineskip)$) -- ++(-\labellength\baselineskip,0pt) -- ++(0pt,-\labelsize\baselineskip);
    \node[anchor=center,mylabel] (#1-beta) at ($(#1.south east)+(0pt,-0.7*\labelsize\baselineskip)+(-0.5*\labellength\baselineskip,0.5*\labelsize\baselineskip)$) { $\beta \approx$ \numprint{#2} };
}
\begin{tikzpicture}[
    inner sep=0pt,
    outer sep=0pt,
    myroundness/.style = {rounded corners=5pt},
    myfig/.style = { shape=rectangle,myroundness,clip },
    myborder/.style = { draw=gray!90!black,thick },
    mylabel/.style = { font=\scriptsize\sf,inner sep=3pt,text opacity=1.0 },
    mylabelbackground/.style = { fill=red,fill opacity=1.0 },
]
    \tikzmath{%
        \numrows = 2;%
        \numcols = 5;%
        \figuresize=0.1925;%
        \labelsize=1.2; % \baselineskip
        \labellength=3.0; % \baselineskip
        \colsep=(1.0-\numcols*\figuresize)/(\numcols-1);%
        \rowsep=\colsep+(\labelsize*0.015);%
    }%
    \node[myfig] (teaser-0-0) { \includegraphics[width=\figuresize\linewidth,]{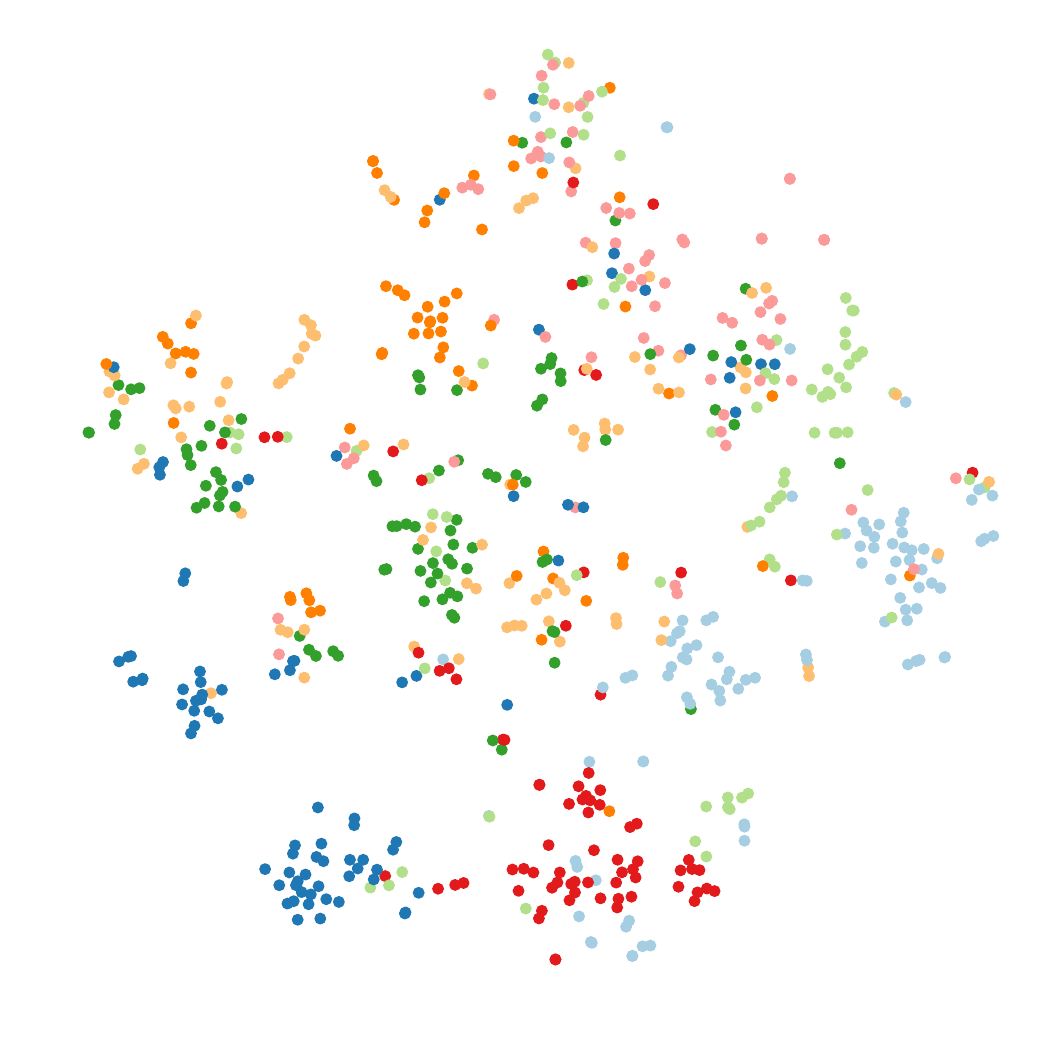} };
    \node[myfig,right=\colsep\linewidth of teaser-0-0] (teaser-1-0) { \includegraphics[width=\figuresize\linewidth,trim={0.2cm 0.2cm 0.2cm 0.2cm},clip]{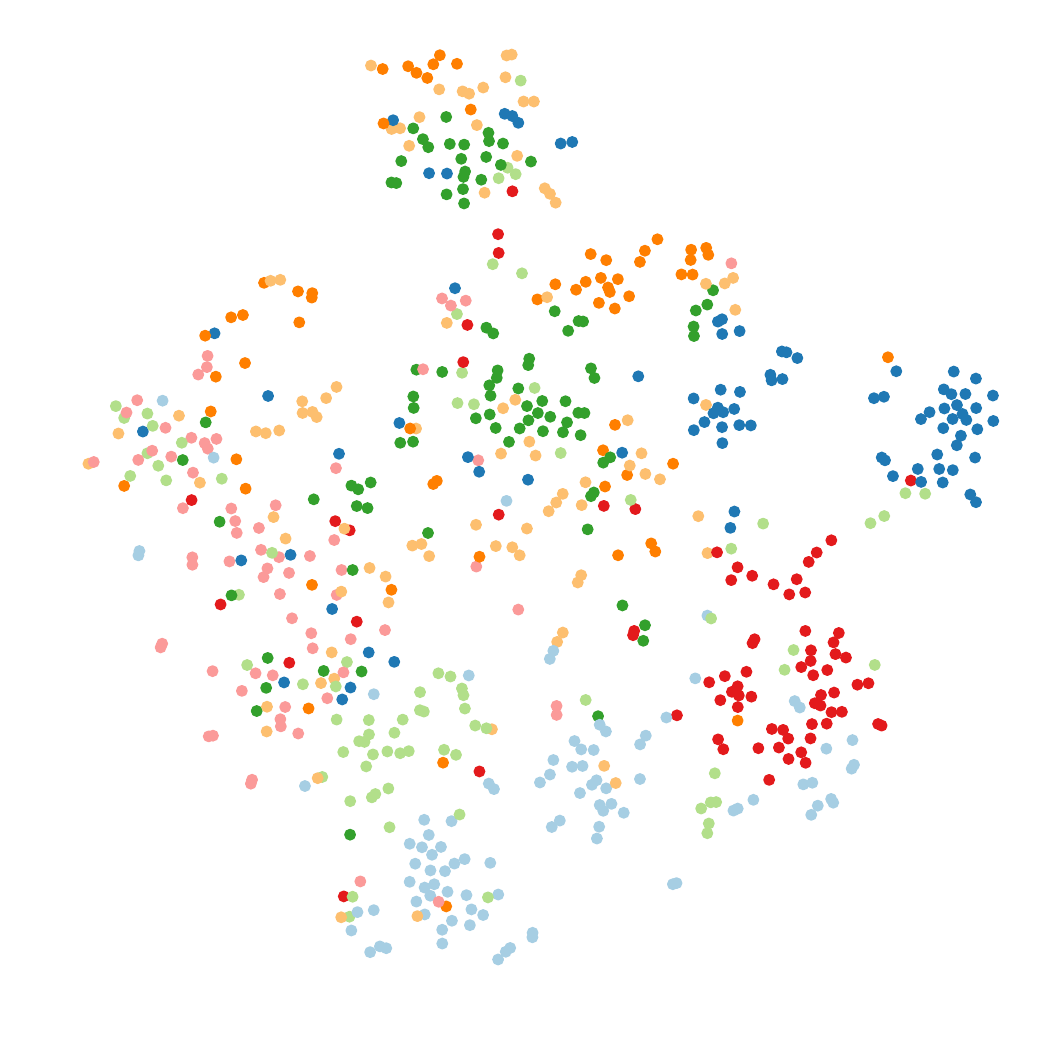} };
    \node[myfig,right=\colsep\linewidth of teaser-1-0] (teaser-2-0) { \includegraphics[width=\figuresize\linewidth,trim={0.2cm 0.2cm 0.2cm 0.2cm},clip]{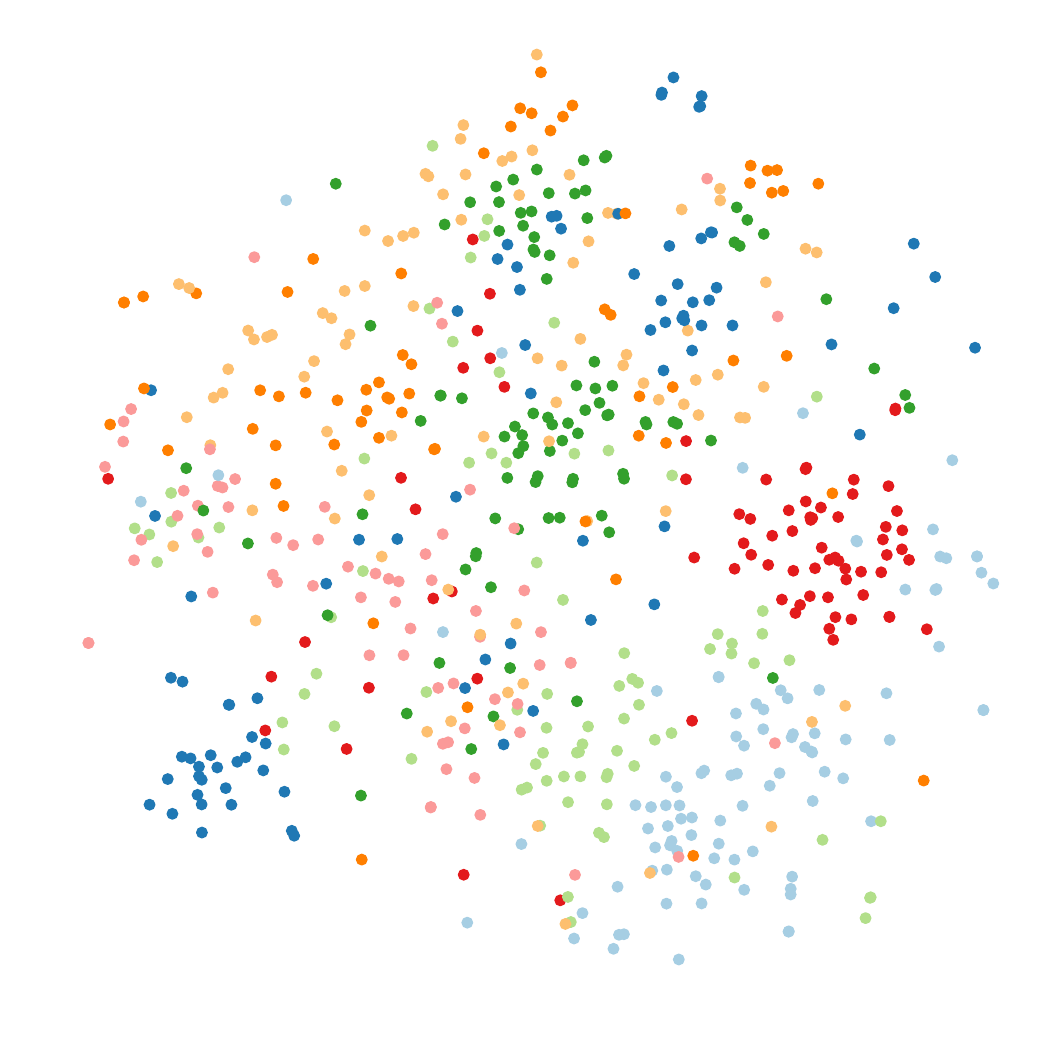} };
    \node[myfig,right=\colsep\linewidth of teaser-2-0] (teaser-3-0) { \includegraphics[width=\figuresize\linewidth,trim={0.2cm 0.2cm 0.2cm 0.2cm},clip]{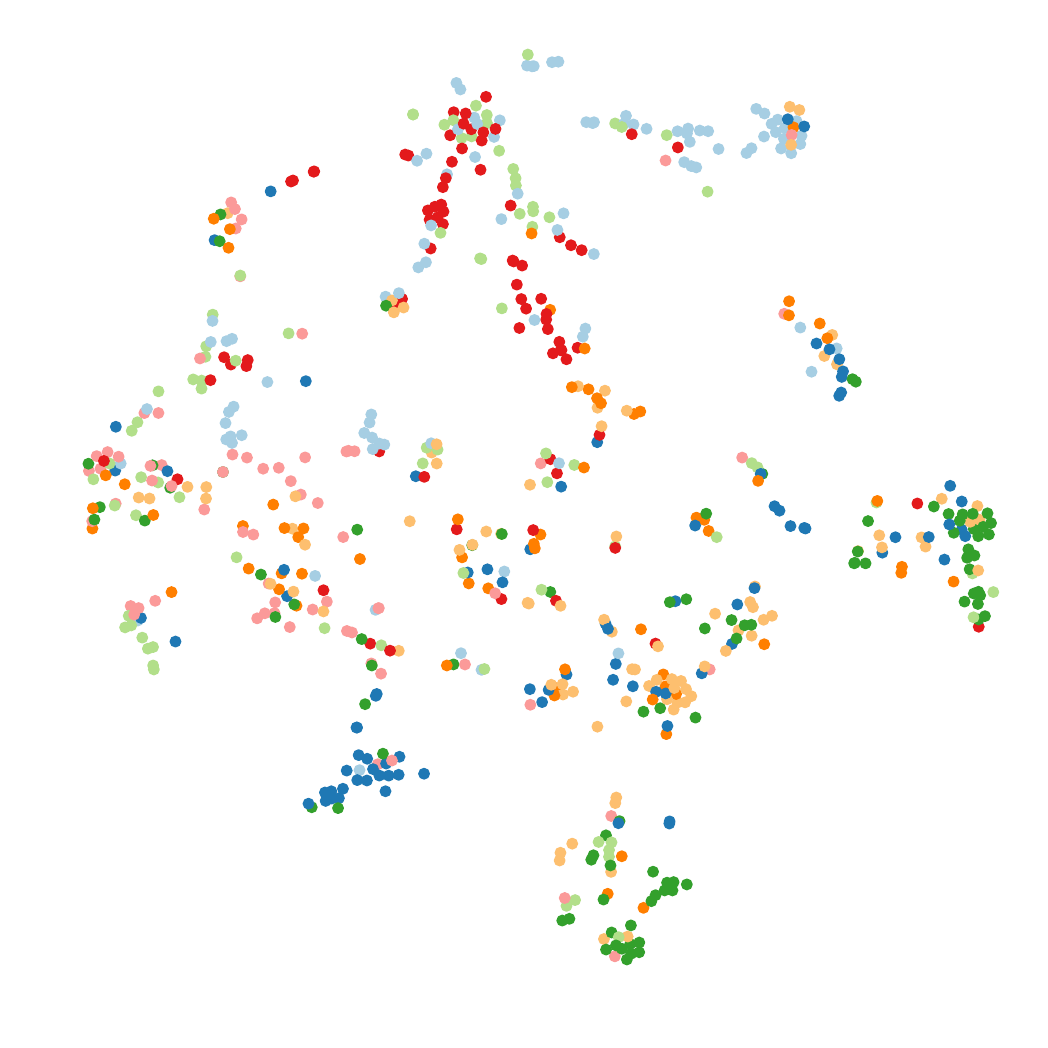} };
    \node[myfig,right=\colsep\linewidth of teaser-3-0] (teaser-4-0) { \includegraphics[width=\figuresize\linewidth,trim={0.2cm 0.2cm 0.2cm 0.2cm},clip]{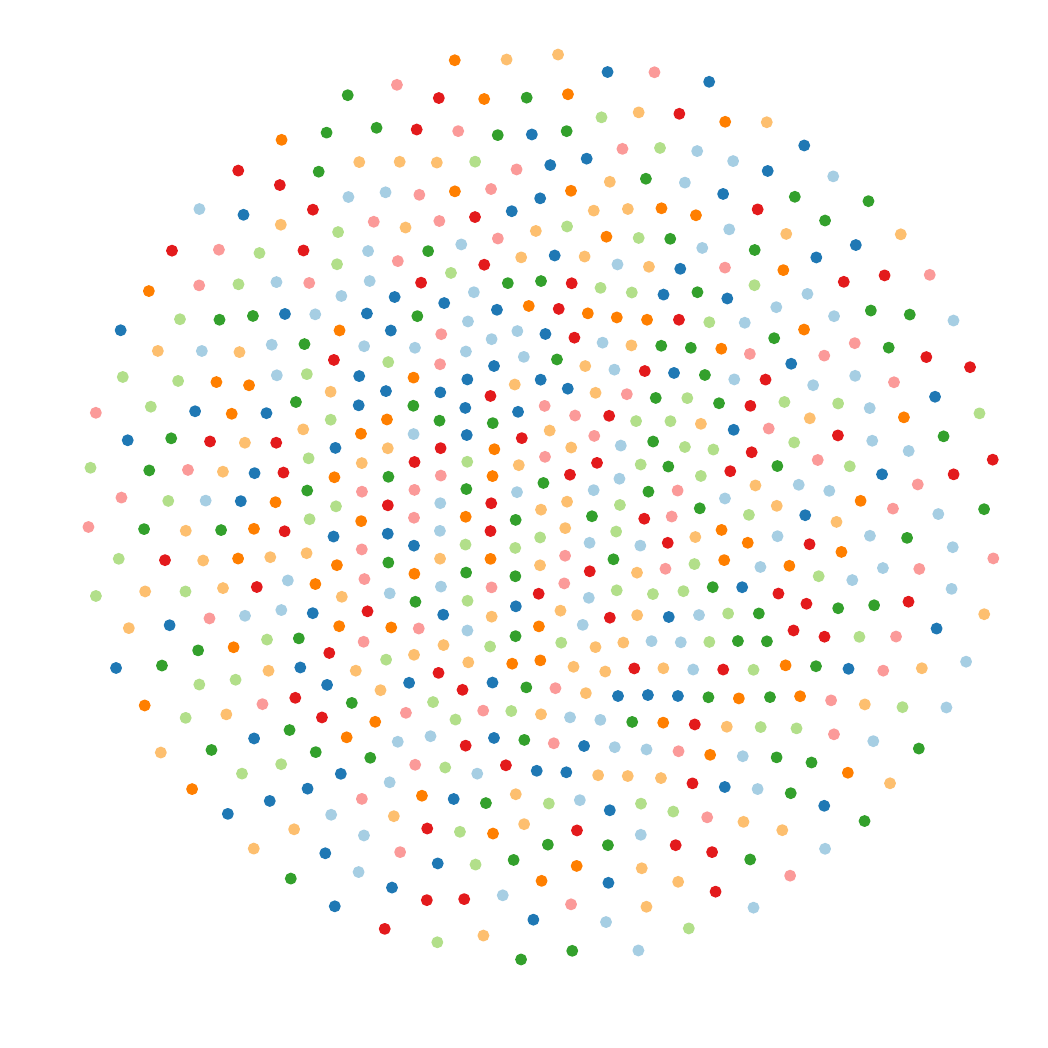} };
    \node[myfig,below=\rowsep\linewidth of teaser-0-0] (teaser-0-1) { \includegraphics[width=\figuresize\linewidth,trim={0.2cm 0.2cm 0.2cm 0.2cm},clip]{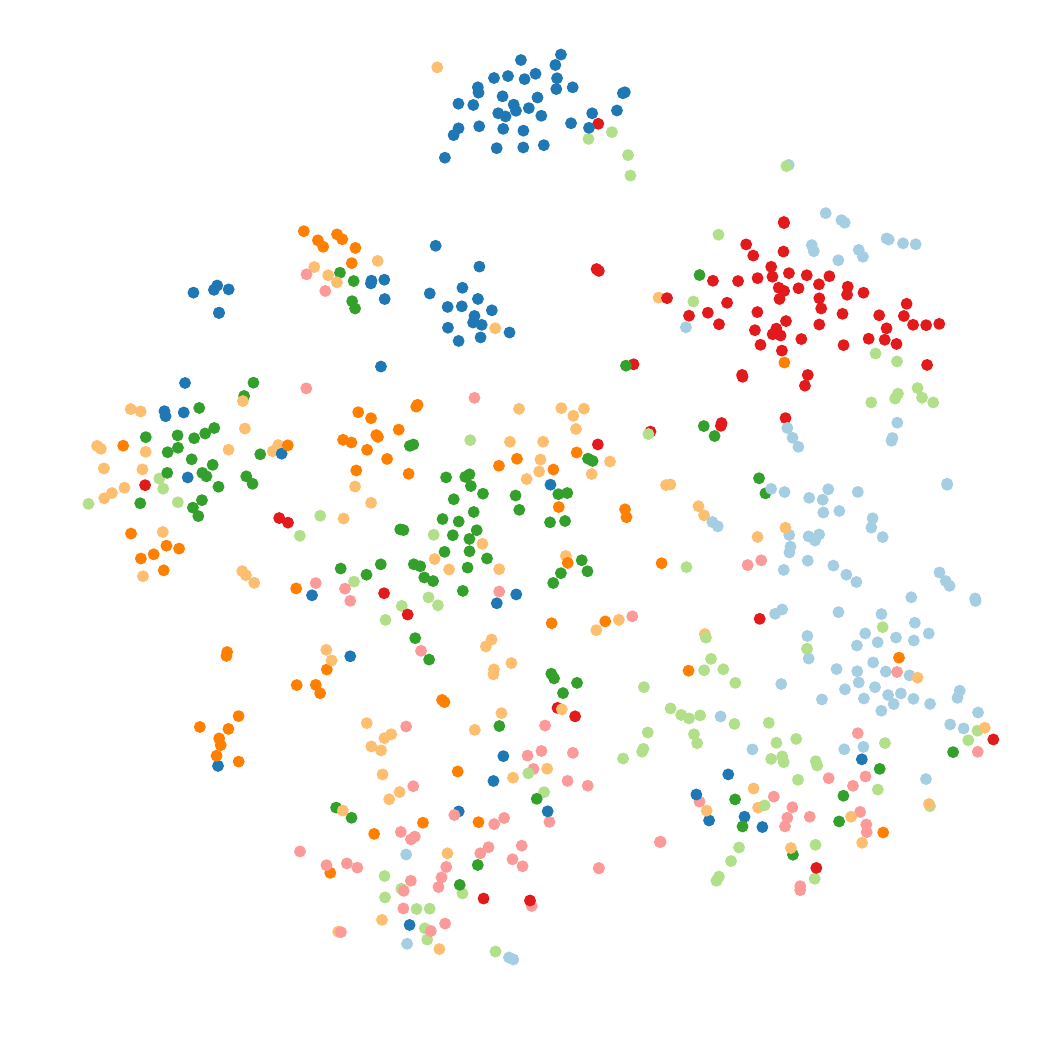} };
    \node[myfig,right=\colsep\linewidth of teaser-0-1] (teaser-1-1) { \includegraphics[width=\figuresize\linewidth,trim={0.2cm 0.2cm 0.2cm 0.2cm},clip]{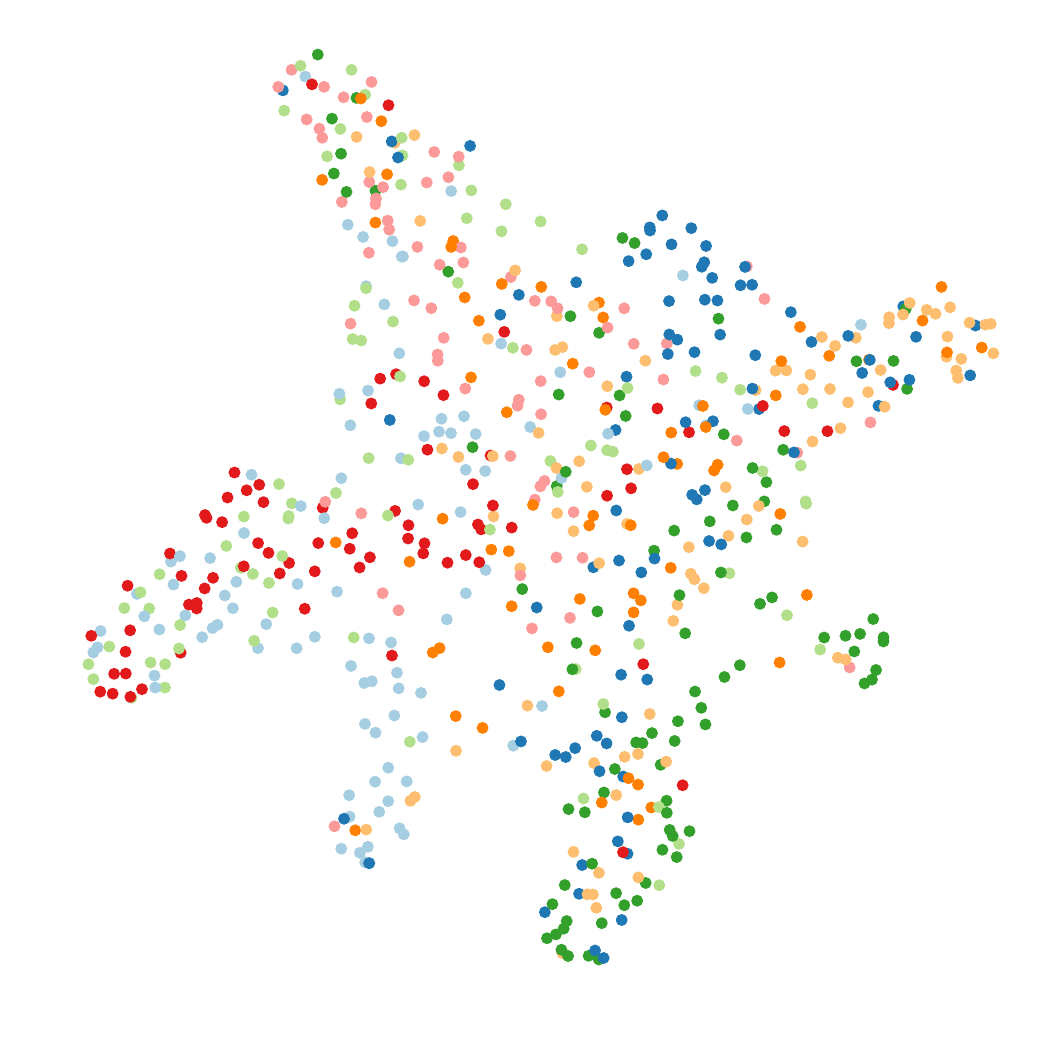} };
    \node[myfig,right=\colsep\linewidth of teaser-1-1] (teaser-2-1) { \includegraphics[width=\figuresize\linewidth,trim={0.2cm 0.2cm 0.2cm 0.2cm},clip]{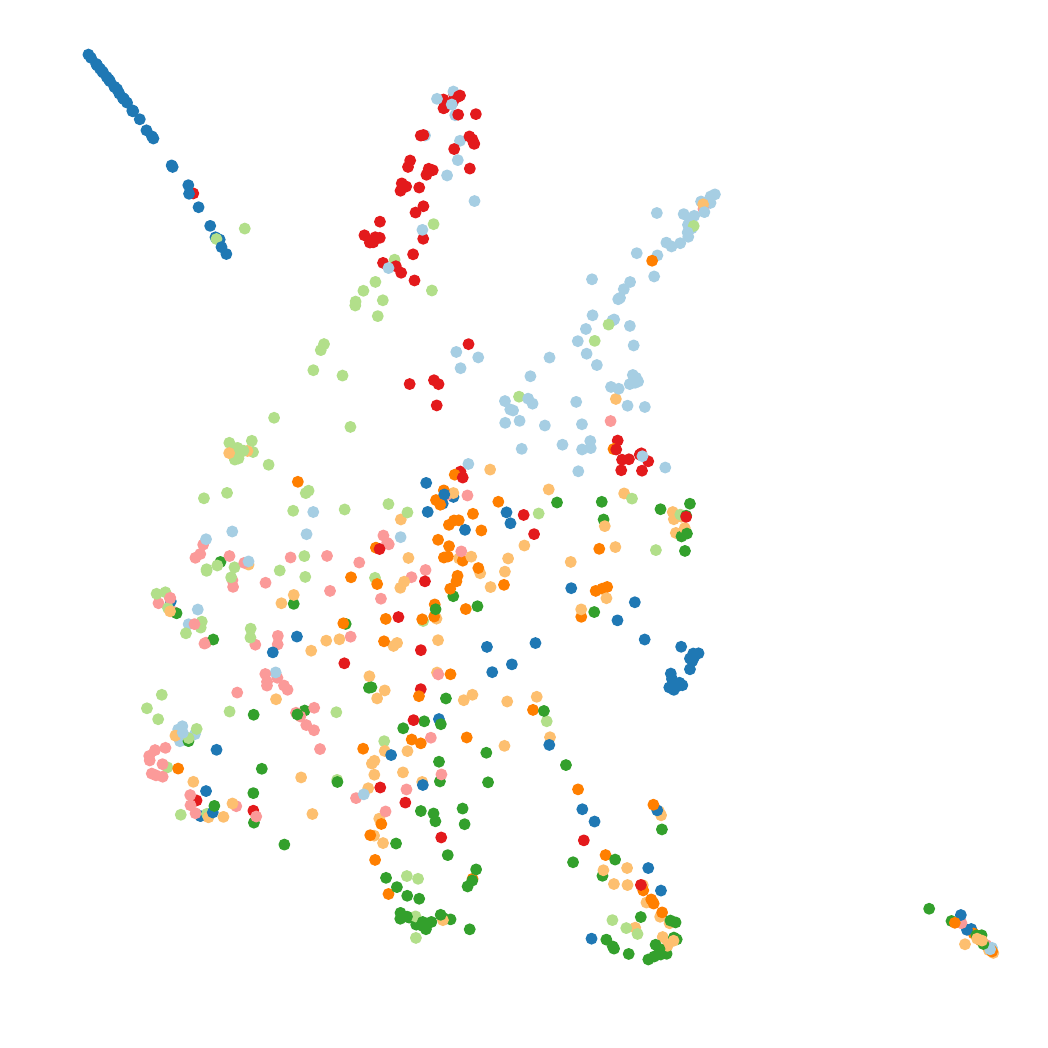} };
    \node[myfig,right=\colsep\linewidth of teaser-2-1] (teaser-3-1) { \includegraphics[width=\figuresize\linewidth,trim={0.2cm 0.2cm 0.2cm 0.2cm},clip]{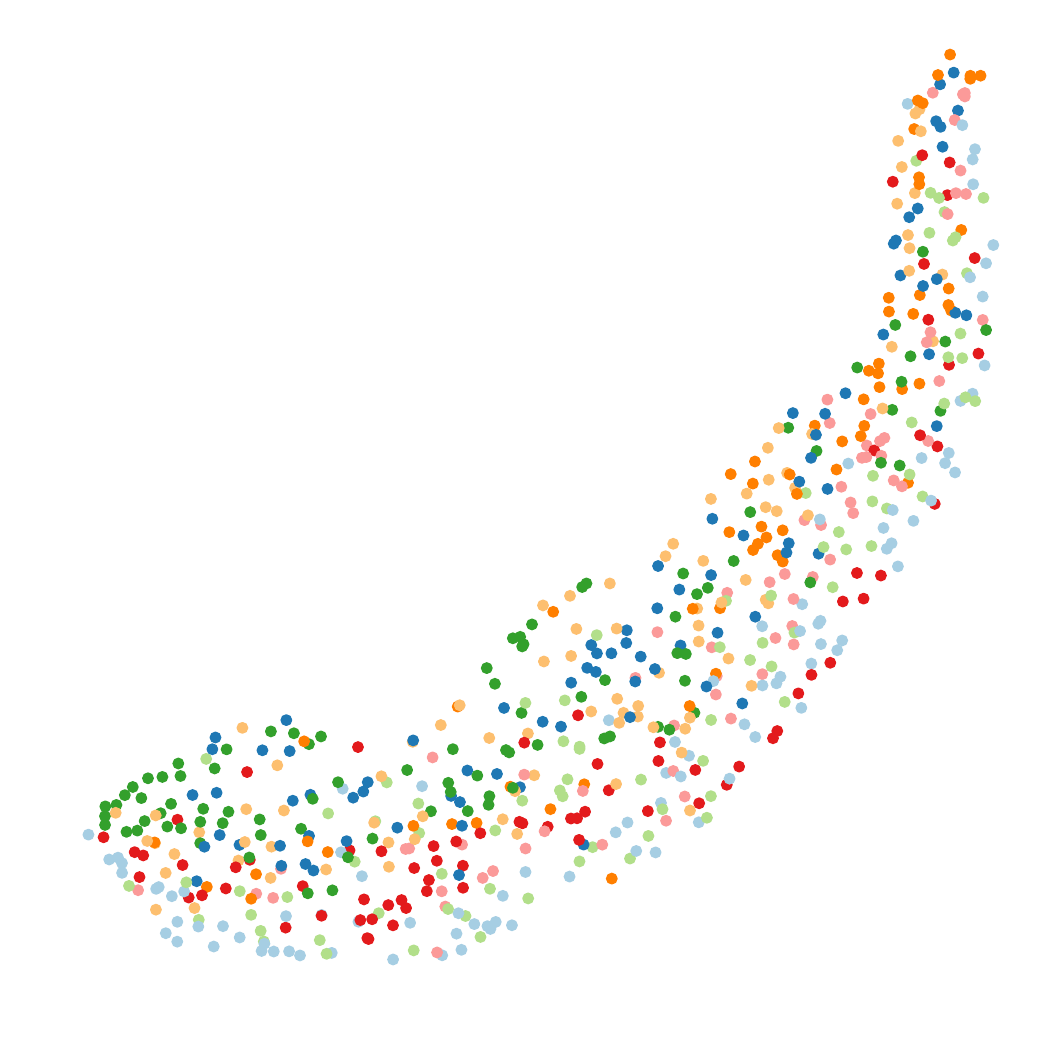} };
    \node[myfig,right=\colsep\linewidth of teaser-3-1] (teaser-4-1) { \includegraphics[width=\figuresize\linewidth,trim={0.2cm 0.2cm 0.2cm 0.2cm},clip]{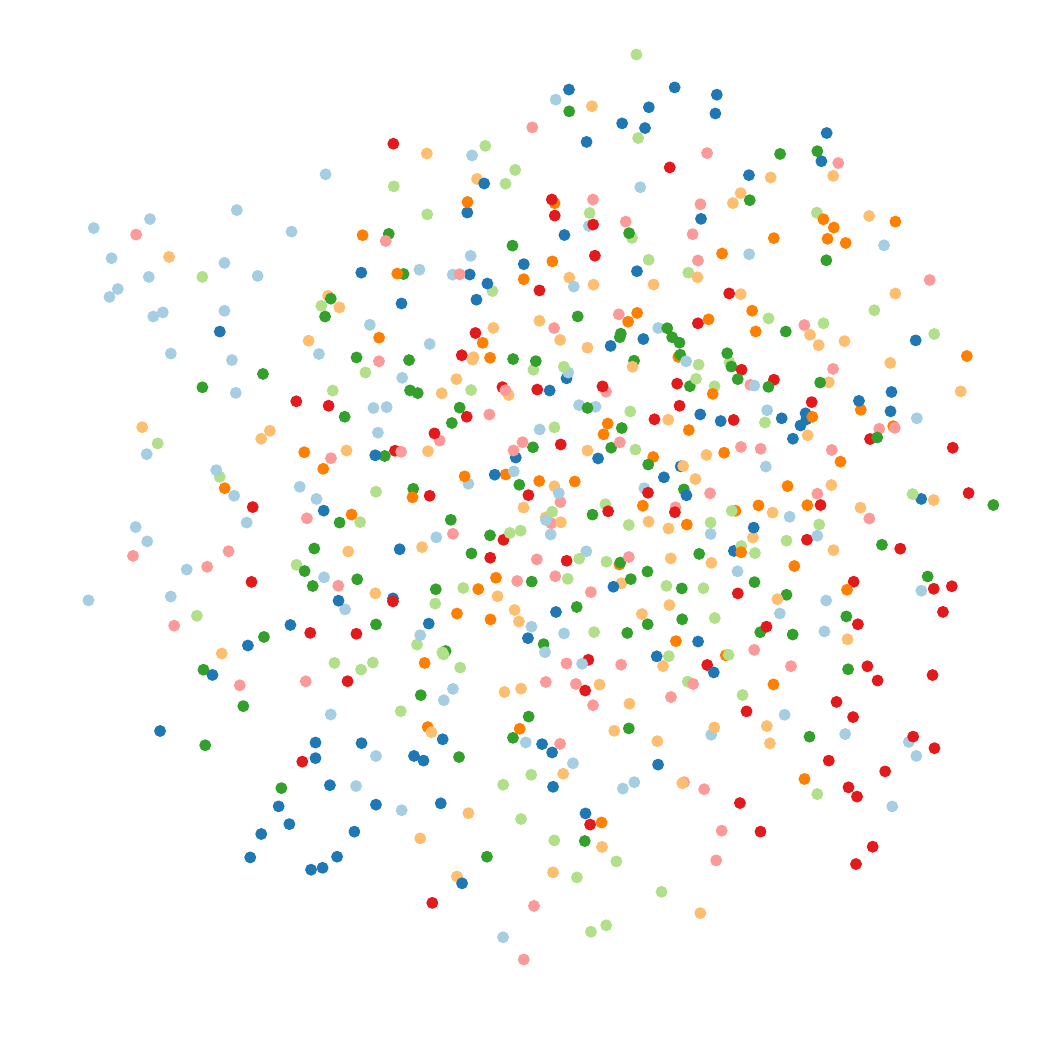} };
    \nprounddigits{2}
    \layoutalphalabel{teaser-0-0}{0.6768427158301402}
    \layoutbetalabel{teaser-0-0}{0.5238740416528698}
    \layoutalphalabel{teaser-1-0}{0.6730083860561258}
    \layoutbetalabel{teaser-1-0}{0.5206232805545947}
    \layoutalphalabel{teaser-2-0}{0.6719129162978279}
    \layoutbetalabel{teaser-2-0}{0.49327051875033456}
    \layoutalphalabel{teaser-3-0}{0.37961334740366454}
    \layoutbetalabel{teaser-3-0}{0.41917531598053415}
    \layoutalphalabel{teaser-4-0}{0.6972733700244266}
    \layoutbetalabel{teaser-4-0}{0.5089650598892596}
    
    \layoutalphalabel{teaser-0-1}{0.672258551200556}
    \layoutbetalabel{teaser-0-1}{0.5136370635108491}
    \layoutalphalabel{teaser-1-1}{0.6465609603238471}
    \layoutbetalabel{teaser-1-1}{0.48554443675780173}
    \layoutalphalabel{teaser-2-1}{0.6966807459875519}
    \layoutbetalabel{teaser-2-1}{0.502266813399254}
    \layoutalphalabel{teaser-3-1}{0.4996461661873002}
    \layoutbetalabel{teaser-3-1}{0.4397327163153816}
    \layoutalphalabel{teaser-4-1}{0.416800217184564}
    \layoutbetalabel{teaser-4-1}{0.43355465096508816}
    \node[myborder,myroundness,fit={(teaser-0-0)(teaser-0-0-alpha)(teaser-0-0-beta)}] {};
    \node[myborder,myroundness,fit={(teaser-1-0)(teaser-1-0-alpha)(teaser-1-0-beta)}] {};
    \node[myborder,myroundness,fit={(teaser-2-0)(teaser-2-0-alpha)(teaser-2-0-beta)}] {};
    \node[myborder,myroundness,fit={(teaser-3-0)(teaser-3-0-alpha)(teaser-3-0-beta)}] {};
    \node[myborder,myroundness,fit={(teaser-4-0)(teaser-4-0-alpha)(teaser-4-0-beta)}] {};
    \node[myborder,myroundness,fit={(teaser-0-1)(teaser-0-1-alpha)(teaser-0-1-beta)}] {};
    \node[myborder,myroundness,fit={(teaser-1-1)(teaser-1-1-alpha)(teaser-1-1-beta)}] {};
    \node[myborder,myroundness,fit={(teaser-2-1)(teaser-2-1-alpha)(teaser-2-1-beta)}] {};
    \node[myborder,myroundness,fit={(teaser-3-1)(teaser-3-1-alpha)(teaser-3-1-beta)}] {};
    \node[myborder,myroundness,fit={(teaser-4-1)(teaser-4-1-alpha)(teaser-4-1-beta)}] {};
\end{tikzpicture}%
    \caption{Two-dimensional scatter plots derived from topic models and a subsequent dimensionality reduction algorithm. Each point represents the merged source code files of a software project from \textit{GitHub}. The color indicates the associated class of the software project, e.g., \textit{shell}, or \textit{frontend}. The values $\alpha$ and $\beta$ correspond to our aggregated accuracy and perception metric, respectively.}
    \label{fig:teaser}
}

\graphicspath{{figs/}{figures/}{pictures/}{images/}{./}} % where to search for the images

\usepackage{mathptmx}                  % use matching math font

\DeclareMathAlphabet{\mathcal}{OMS}{cmsy}{m}{n}
\usepackage{numprint}
\npdecimalsign{.} % Default is , (German version)
\usepackage{booktabs}
\usepackage{tabularx}
\usepackage{csquotes}
\usepackage{amsfonts}
\usepackage{amssymb}
\usepackage{extarrows}
\usepackage[export]{adjustbox}
\usepackage{tikz}
\usetikzlibrary{math,calc,positioning,fit,arrows.meta}
\usepackage{makecell}
\usepackage{fontawesome}

\usepackage{booktabs}
\usepackage{multirow}
\usepackage{diagbox}
\usepackage{collcell}
\usepackage{pgfmath}
\usepackage{pgfplots}
\usepackage{pgfplotstable}

\usepackage{pdfpages}

\usepackage{xcolor}
\usepackage{pifont}% http://ctan.org/pkg/pifont
\newcommand{\yes}{\textcolor{black}{\ding{51}}} % checkmark
\newcommand{\no}{\textcolor{black}{\ding{55}}}% crossmark

\newcommand{\revisedcontent}[1]{#1}

\definecolor{YlOrBr-A}{HTML}{ffffd4}
\definecolor{YlOrBr-B}{HTML}{fed98e}
\definecolor{YlOrBr-C}{HTML}{fe9929}
\definecolor{YlOrBr-D}{HTML}{d95f0e}
\definecolor{YlOrBr-E}{HTML}{993404}

\pgfplotsset{%
    colormap={YlOrBr}{color=(YlOrBr-A) color=(YlOrBr-B) color=(YlOrBr-C) color=(YlOrBr-D) color=(YlOrBr-E)},%
    colormap/YlOrBr/.style={
        colormap name=YlOrBr,
    },
}%

\definecolor{Blues-A}{RGB}{247,251,255}
\definecolor{Blues-B}{RGB}{222,235,247}
\definecolor{Blues-C}{RGB}{198,219,239}
\definecolor{Blues-D}{RGB}{158,202,225}
\definecolor{Blues-E}{RGB}{107,174,214}
\definecolor{Blues-F}{RGB}{66,146,198}
\definecolor{Blues-G}{RGB}{33,113,181}
\definecolor{Blues-H}{RGB}{8,81,156}
\definecolor{Blues-I}{RGB}{8,48,107}

\definecolor{RdBu-A}{RGB}{178,24,43}
\definecolor{RdBu-B}{RGB}{214,96,77}
\definecolor{RdBu-C}{RGB}{244,165,130}
\definecolor{RdBu-D}{RGB}{253,219,199}
\definecolor{RdBu-E}{RGB}{247,247,247}
\definecolor{RdBu-F}{RGB}{209,229,240}
\definecolor{RdBu-G}{RGB}{146,197,222}
\definecolor{RdBu-H}{RGB}{67,147,195}
\definecolor{RdBu-I}{RGB}{33,102,172}

\definecolor{Reds-A}{RGB}{255,245,240}
\definecolor{Reds-B}{RGB}{254,224,210}
\definecolor{Reds-C}{RGB}{252,187,161}
\definecolor{Reds-D}{RGB}{252,146,114}
\definecolor{Reds-E}{RGB}{251,106,74}
\definecolor{Reds-F}{RGB}{239,59,44}
\definecolor{Reds-G}{RGB}{203,24,29}
\definecolor{Reds-H}{RGB}{165,15,21}
\definecolor{Reds-I}{RGB}{103,0,13}

\pgfplotsset{%
    colormap={Blues}{color=(Blues-A) color=(Blues-B) color=(Blues-C) color=(Blues-D) color=(Blues-E) color=(Blues-F) color=(Blues-G)},% color=(Blues-H) color=(Blues-I) % limit to 7 colors
    colormap/Blues/.style={
        colormap name=Blues,
    },
}%

\pgfplotsset{%
    colormap={RdBu}{color=(RdBu-A) color=(RdBu-B) color=(RdBu-C) color=(RdBu-D) color=(RdBu-E) color=(RdBu-F) color=(RdBu-G) color=(RdBu-H) color=(RdBu-I) },
    colormap/RdBu/.style={
        colormap name=RdBu,
    },
}%

\pgfplotsset{%
    colormap={Reds}{color=(Reds-A) color=(Reds-B) color=(Reds-C) color=(Reds-D) color=(Reds-E) color=(Reds-F) color=(Reds-G) color=(Reds-H) color=(Reds-I) },
    colormap/Reds/.style={
        colormap name=Reds,
    },
}%

\newcommand{\CalcCellColor}[1]{%
    \pgfmathfloatparsenumber{#1}%
    \pgfmathfloattofixed{\pgfmathresult}%
    \let\value=\pgfmathresult
    \pgfplotscolormapaccess[0:7]{\value}{Blues}%
    \xdef\temp{%
        \noexpand\cellcolor[rgb]{\pgfmathresult}%
    }%
    \temp%
    {#1}
}%

\newcommand{\CalcCellColorP}[1]{%
    \pgfmathfloatparsenumber{#1}%
    \pgfmathfloattofixed{\pgfmathresult}%
    \let\value=\pgfmathresult
    \pgfplotscolormapaccess[0:100]{\value}{Blues}%
    \xdef\temp{%
        \noexpand\cellcolor[rgb]{\pgfmathresult}%
    }%
    \temp%
    {%
        \nprounddigits{0}%
        \scriptsize\npmakebox[99\,\%][r]{\numprint[\%]{#1}}%
    }%
}%

\newcolumntype{R}{>{\collectcell{\CalcCellColor}}c<{\endcollectcell}}
\newcolumntype{P}{>{\collectcell{\CalcCellColorP}}c<{\endcollectcell}}

\definecolor{RdYlBu-A}{HTML}{4575b4}
\definecolor{RdYlBu-B}{HTML}{91bfdb}
\definecolor{RdYlBu-C}{HTML}{e0f3f8}
\definecolor{RdYlBu-D}{HTML}{ffffbf}
\definecolor{RdYlBu-E}{HTML}{fee090}
\definecolor{RdYlBu-F}{HTML}{fc8d59}
\definecolor{RdYlBu-G}{HTML}{d73027}

\definecolor{YlOrRd-A}{HTML}{ffffb2}
\definecolor{YlOrRd-B}{HTML}{fed976}
\definecolor{YlOrRd-C}{HTML}{feb24c}
\definecolor{YlOrRd-D}{HTML}{fd8d3c}
\definecolor{YlOrRd-E}{HTML}{fc4e2a}
\definecolor{YlOrRd-F}{HTML}{e31a1c}
\definecolor{YlOrRd-G}{HTML}{b10026}

\pgfplotsset{%
    colormap={RdYlBu-CM}{color=(RdYlBu-A) color=(RdYlBu-B) color=(RdYlBu-C) color=(RdYlBu-D) color=(RdYlBu-E) color=(RdYlBu-F) color=(RdYlBu-G)},%
    colormap/RdYlBu-CM/.style={
        colormap name=RdYlBu-CM,
    },
}%

\pgfplotsset{%
    colormap={YlOrRd-CM}{color=(YlOrRd-A) color=(YlOrRd-B) color=(YlOrRd-C) color=(YlOrRd-D) color=(YlOrRd-E) color=(YlOrRd-F) color=(YlOrRd-G)},%
    colormap/YlOrRd-CM/.style={
        colormap name=YlOrRd-CM,
    },
}%

\pgfplotsset{%
    colormap={YlOrRd-CM-inv}{color=(YlOrRd-G) color=(YlOrRd-F) color=(YlOrRd-E) color=(YlOrRd-D) color=(YlOrRd-C) color=(YlOrRd-B) color=(YlOrRd-A)},%
    colormap/YlOrRd-CM-inv/.style={
        colormap name=YlOrRd-CM-inv,
    },
}%

\pgfplotstableset{
    /color cells/min/.initial=0,
    /color cells/max/.initial=1000,
    /color cells/textcolor/.initial=,
    color cells/.code={%
        \pgfqkeys{/color cells}{#1}%
        \pgfkeysalso{%
            postproc cell content/.code={%
                \begingroup
                \pgfkeysgetvalue{/pgfplots/table/@preprocessed
                  cell content}\value
                \ifx\value\empty
                    \endgroup
                \else
                \pgfmathfloatparsenumber{\value}%
                \pgfmathfloattofixed{\pgfmathresult}%
                \let\value=\pgfmathresult
                \pgfplotscolormapaccess
                    [\pgfkeysvalueof{/color cells/min}:\pgfkeysvalueof{/color
                      cells/max}]
                    {\value}
                    {\pgfkeysvalueof{/pgfplots/colormap name}}%
                \pgfkeysgetvalue{/pgfplots/table/@cell content}\typesetvalue
                \pgfkeysgetvalue{/color cells/textcolor}\textcolorvalue
                \toks0=\expandafter{\typesetvalue}%
                \xdef\temp{%
                    \noexpand\pgfkeysalso{%
                        @cell content={%
                            \noexpand\cellcolor[rgb]{\pgfmathresult}%
                            \noexpand\definecolor{mapped
                              color}{rgb}{\pgfmathresult}%
                            \ifx\textcolorvalue\empty
                            \else
                                \noexpand\color{\textcolorvalue}%
                            \fi
                            \the\toks0 %
                        }%
                    }%
                }%
                \endgroup
                \temp
                \fi
            }%
        }%
    }
}

\pgfplotstableset{
    colorbarstyle/.style={
        numeric type,
        fixed,
        fixed zerofill,
        precision=1,
        skip 0.=false,
        color cells={min=0,max=1},
        column type=r,
        column type/.add={>{\footnotesize}}{},
        /pgfplots/colormap name=RdYlBu-CM,
    }
}

\begin{document}

%%%%%%%%%%%%%%%%%%%%%%%%%%%%%%%%%%%%%%%%%%%%%%%%%%%%%%%%%%%%%%%%
%%%%%%%%%%%%%%%%%%%%%% START OF THE PAPER %%%%%%%%%%%%%%%%%%%%%%
%%%%%%%%%%%%%%%%%%%%%%%%%%%%%%%%%%%%%%%%%%%%%%%%%%%%%%%%%%%%%%%%

%\firstsection{Introduction}

\maketitle

\section{Introduction}

Text data is generated in large quantities in numerous application domains, e.g., social media, news articles, scientific articles, and literature.
Given a set of documents, a so-called corpus, numerous text visualizations have been proposed to support users in various analytic tasks, e.g., summarization, sentiment analysis, or exploration~\cite{kucher2019TextVisualizationRevisited}.
In order to \enquote{leverage the cognitive benefits from cartography as an established body of knowledge for information visualization}, many text visualization techniques rely on a map-like metaphor, i.e., \enquote{a map imitation that makes spatialized data appear more like a cartographic map by emphasizing spatial context}~\cite{hografer2020state}.
To derive a spatialization, which \enquote{encode[s] similarities, [by] mapping each data item to a point on the visual space such that the relative pairwise proximities reflect at best the corresponding pairwise similarities}~\cite{Nonato2019MDP}, different \textit{Topic Models} (TMs) and \textit{Dimensionality Reductions} (DRs) are applied for generating two-dimensional layouts for a corpus.
Starting from the \textit{Document-Term Matrix} (DTM) representation of a corpus, which stores the frequency of the terms within each document, TMs aim to detect clusters within the vocabulary, so-called topics, by analyzing patterns of co-occurring words~\cite{crain2012dimensionality}.
Thereby, TMs are specified by several hyperparameters that control the underlying model and the training algorithm.
As a result, TMs yield a high-dimensional vector representation of the documents, which are further projected to a lower-dimensional space by a DR for visualization.
\revisedcontent{\Cref{tab:vis techniques} lists examples of text visualization techniques using a two-dimensional spatialization derived from a TM and a DR.}
\revisedcontent{To summarize, starting from a high-dimensional description of the documents within a corpus that stores the term frequencies, combining a topic model and a subsequent dimensionality reduction results in a two-dimensional scatter plot, where each point represents a single document.}

\begin{table}[t]
    %\tiny
    \footnotesize
    \centering
	\setlength{\tabcolsep}{4.25pt}%
	\renewcommand{\arraystretch}{1.00}%
    \caption{Examples of text visualizations that rely on a two-dimensional layout derived from a TM and a DR.
    }
    \vspace{-0.5\baselineskip}
    \begin{tabular}{rcccc}
    \toprule
    \multicolumn{1}{c}{\textbf{Visualization}} & \textbf{TM} & \textbf{tf-idf} & \textbf{DR} & \textbf{linear combination} \\ \midrule
    Skupin \cite{skupin2004worldofgeography} & VSM & \no & SOM & \no \\
    Kuhn et al. \cite{kuhn2010software} & LSI & \no & MDS & \no \\
    Fried et al. \cite{Fried2014MapsComputerScience} & LSI & \yes & MDS & \no \\
    Caillou et al. \cite{Caillou2021Cartolabe} & LSI & \yes & UMAP & \no \\
    Kim et al. \cite{Kim2017TopicLens} & NMF & \no & t-SNE & \no \\
    Choo et al. \cite{Choo2013Utopian} & NMF & \no & t-SNE & \no  \\
    Linstead et al. \cite{linstead09sourcerer} & LDA & \no & MDS & \no \\
    Gansner et al. \cite{gansner2013interactive} & LDA & \no & MDS & \no  \\
    Atzberger et al. \cite{atzberger2021softwareforest,atzberger2022-knowhowmap} & LDA & \no & MDS & \yes \\
    Kucher et al. \cite{kucher2018AnalysisVinciProceedings} & LDA & \no & t-SNE & \no \\
    Yan et al. \cite{Yan2019InteractiveVisualAnalytics} & LDA & \no & t-SNE & \no \\
    \bottomrule
    \end{tabular}
    \label{tab:vis techniques}
\end{table}

A visualization designer has to select a TM and a DR and specify the corresponding hyperparameters.
\Cref{fig:teaser} shows exemplarily that the concrete choice of the combination and assignment of the hyperparameters strongly influences the resulting layout regarding clustering and global structure.
TMs and DRs are often treated as a \enquote{black box} without thoroughly investigating the effect of the hyperparameters.
In most cases, publications presenting a text visualization focus on the visual mapping, whereas a comparison of several possible layouts is usually omitted.
Other studies that focus on deriving guidelines for the effective use of DRs for visualization tasks have not yet considered TMs.
We close this gap by investigating the influence of TMs and DRs on the spatialization of text visualizations.
The research space of TM, DR, and test data set is extensive.
We narrow down this large experimental space by means of a computing cluster and a representative selection of relevant methods.
This is, to the best of our knowledge, the most comprehensible experimental evaluation in this area.

We present a benchmark $\mathcal{B} = (\mathcal{D}, \mathcal{L}, \mathcal{Q})$ composed of a set of corpora $\mathcal{D}$, layout algorithms $\mathcal{L}$, and quality metrics $\mathcal{Q}$.
The set $\mathcal{D}$ contains five corpora, each given by a DTM, with assigned class labels for each document.
Each layout in $\mathcal{L}$ emerges from a combination of a TM and a subsequent DR.
The set $\mathcal{L}$ comprises 52 different layout algorithms composed of 13 TMs and 4 DRs.
The metrics in $\mathcal{Q}$ quantify the preservation of local and global structures of the high-dimensional corpora in the two-dimensional layout and their cluster separation.
We quantify each layout and the influence of their hyperparameters by performing a grid search using a computational cluster.
We derived a tabular dataset from the execution of the benchmark with more than \numprint{45000} sampled layouts and their quality scores.
Based on the dataset, we show the performances with respect to accuracy and the effectiveness to perceive clusters and propose guidelines for the effective use of TMs and DRs for the spatialization of corpora.
To summarize, we make the following contributions:
\begin{enumerate}
    \setlength{\itemsep}{0pt}
    \item We provide a benchmark of five corpora, 52 layout algorithms, and eight quality metrics for evaluating the use of TMs and DRs for two-dimensional text spatialization.
    We provide our implementation as a Git repository\footnote{\href{https://github.com/cgshpi/Topic-Models-and-Dimensionality-Reduction-Benchmark}{\revisedcontent{\faGithub{} hpicgs/Topic-Models-and-Dimensionality-Reduction-Benchmark}}}.% benchmark
    \item We generate a large multivariate dataset with more than \numprint{45000} entries containing the quality score of different hyperparameter configurations.
    We selected the range of the hyperparameters following the recommendations of the used libraries.
    \item We provide high-level insights into the aggregated performances and propose guidelines for the practical use of TMs and DRs.
\end{enumerate}
The remaining part is structured as follows:
we present related work in \Cref{section: related work}.
The structure and composition of our benchmark, together with technical details on the execution, are detailed in \Cref{section: benchmark}.
In \Cref{section: results}, we analyze the resulting multivariate dataset.
We discuss the results and present guidelines together with threats to validity in \Cref{section: discussion}.
We conclude this work and present directions for future work in \Cref{section: conclusions}.

\section{Related Work}
\label{section: related work}

Several text analysis tasks, e.g., text classification~\cite{aggarwal2012surveyClassification}, text summarization~\cite{nenkova2012surveySummarization}, or text clustering~\cite{aggarwal2012surveyClustering}, rely on TMs for modeling documents within a corpus.
In most cases, TMs are evaluated using statistical measures, e.g., perplexity or coherence scores~\cite{Roeder2015coherence}, or by asking users about the interpretability of the derived topics~\cite{Lipton2018Interpretability}.
Although TMs are often part of text visualizations or intended to be represented by a visualization, e.g., for topic comparison~\cite{Alexander2016ComparisonTopicModels,Xu2013TopicCompetition,sievert2014ldavis}, for topic evolution~\cite{dang2019wordstream}, or corpus exploration~\cite{Dou2013HierarchicalTopics,Peter2015Topicks}, in most cases TMs are not evaluated.
An exception is the work of Riehmann et al., who showed in an expert study that the topics extracted by LDA do not match the list that experts curated~\cite{Riehmann2019VisualizingThinkersLife}.
DRs are among the popular techniques for visualizing high-dimensional data, e.g., by computing a set of scatter plots~\cite{Lehmann2016OptimalProjections}, or by computing two-dimensional layouts as shown in \revisedcontent{\Cref{tab:vis techniques}}.
Thereby, DRs are widely evaluated and discussed in previous surveys with differing foci: surveys that focus on the mathematical principles of DRs and quantitative studies.
In the latter, either the accuracy of DRs, i.e., the preservation of local and global structures, or the effectiveness of the resulting layout for human perception are investigated.

\subsection{Mathematical Surveys of Dimensionality Reductions}

One of the earlier surveys on DRs with a mathematical introduction was presented by Fodor~\cite{fodor2002survey}.
Using a similar approach, Cunningham and Ghahramani discussed several linear DRs~\cite{cunningham2015linear}.
Further, Engel et al.\ presented a survey of basic DRs from a visualization point of view~\cite{engel2012survey}.
In addition to the theoretical alignment of the individual DRs, the authors also compare their underlying assumptions, online compatibility, and computational cost.
Nonato and Aupetit recently presented a comprehensive survey of DRs specific to the visualization domain~\cite{Nonato2019MDP}.
In addition to a detailed taxonomy of DRs, distortion types, analytics tasks, and layout enrichment methods, the authors formulate guidelines on choosing a DR for a given analytics task.
Unlike our work, the authors \revisedcontent{additionaly} formulate guidelines derived from their proposed taxonomy and the constraints on mathematical properties of the DRs, rather than from experimental results.

\subsection{Evaluating Dimensionality Reductions for Accuracy}
One method for assessing visualization techniques is to define aspects of quality and derive quality metrics for quantitative measurements~\cite{behrisch2018quality}.
To derive guidelines for the effective use of DRs for visualization techniques, benchmarks that evaluate a large number of different layouts have been proposed.
One quality aspect for DRs is the accuracy, i.e., the preservation of local and global structures from the high-dimensional dataset in the low-dimensional representation, for which different metrics have been developed~\cite{Morariu2023PredictingUserPreferences}.
For example, van der Maaten et al.\ proposed a benchmark for comparing different DRs~\cite{van2009dimensionality}.
Their study compared the quality of twelve non-linear methods with Principal Component Analysis (PCA) on ten different datasets by measuring three quality metrics.
As a main result, the authors showed that the non-linear methods did not outperform PCA.
However, t-SNE and UMAP were not included in the benchmark.
One study by Gisbrecht and Hammer presents the mathematical principles of non-linear DRs and an evaluation of their performance on three datasets~\cite{gisbrecht2015data}.
In particular, the authors investigate the influence of individual hyperparameters on the quality of the resulting layouts.
Recently, Espadoto et al.\ presented an architecture for a DR benchmark to support users in selecting appropriate DRs and allow researchers to assess new methods~\cite{espadoto2020selecting}.
Further, the authors evaluated the benchmark to derive insights and best practices for the effective use of DRs for visualization tasks~\cite{Espadoto2021Toward}.
Their benchmark comprises 18 datasets, 44 dimensionality reductions, and seven quality metrics.
Although the authors consider text data as one of three types besides tabular and image data, their benchmark does not consider TMs.
Similarly, Vernier et al.\ followed this approach and investigated which DRs are suitable for visualizing dynamic data~\cite{vernier2020quantitative}.
For this purpose, the benchmark was extended by metrics that measure the temporal stability of the generated layouts.
Until then, the assessment of temporal stability was approached by human judgement~\cite{garcia2013stability}.
In a later work, Vernier et al. extended their benchmark by two versions of t-SNE that improved spatial quality and temporal stability~\cite{vernier2021guided}.

\subsection{Evaluating Dimensionality Reductions for Perception}
Complimentary to accuracy metrics, perception metrics quantify the effectiveness of a two-dimensional layout for perceiving structures, e.g., class separation~\cite{sedlmair2015data,Albuquerque2011PerceptionBased}.
Morariu et al. presented a benchmark for investigating how quality metrics are suitable to describe the visual appearance of two-dimensional layouts derived from DRs~\cite{Morariu2023PredictingUserPreferences}.
For this purpose, rankings provided by study participants were used as labels to predict user preferences given quality metrics.
Similarly, Xia et al.\ presented a convolutional neural network for modeling the human perception of visual clusters~\cite{Xia2021VisualClustering}.
Their model is trained on a human-labeled dataset and a qualitative study determining influence factors for cluster perception.
Using a similar approach, Wang et al.\ combined quantitative measurements and human judgments to evaluate their proposed perception-driven DR to maximize the perceived class separation~\cite{Wang2018SupervisedDR}.
Xia et al. presented a contrastive DR approach that considers accuracy metrics in addition to optimizing visual cluster separation by measuring three perception metrics~\cite{Xia2023InteractiveVisualCluster}.
In addition to a quantitative assessment through metrics, Xia et al. showed, through a user study, that their approach outperformed t-SNE and UMAP in the task of cluster identification.
A further work that relies on a user study was presented by Sedlmair et al.\, who investigated to what extent 3D scatter plots or scatter plot matrices improve the perception of cluster separation, compared to 2D scatter plots~\cite{Sedlmair2013EmpiricalGuidance}.
Similarly, Xia et al.\ conducted a user study to investigate which DRs are suitable for visual cluster analysis tasks, e.g., cluster identification, membership identification, distance comparison, and density comparison~\cite{Xia2022RevisitingDimensionalityReduction}.

\section{\texorpdfstring{Benchmark $\mathcal{B}$}{Benchmark}}
\label{section: benchmark}

\begin{table}[t]
    %\tiny
    \footnotesize%
    \centering%
	\setlength{\tabcolsep}{7.0pt}%
	\renewcommand{\arraystretch}{1.00}%
    \vspace{-0.5\baselineskip}
    \caption{Characteristics for the five datasets in our benchmark containing the number of documents $m$, \revisedcontent{the median size of the documents $N$, }the size of the vocabulary $n$ and the number of categories $k$. The size represents the size of the raw dataset given by $m \cdot n \cdot 8$ byte. This measurement neglects program overhead and memory required by the DR and TM.}%
    \begin{tabular}{rrrrrr}
    \toprule
    \multicolumn{1}{c}{\textbf{Dataset}} & \multicolumn{1}{c}{\textbf{Size}} & \multicolumn{1}{c}{\textbf{$m$}} & \multicolumn{1}{c}{\revisedcontent{\textbf{$N$}}} &\multicolumn{1}{c}{\textbf{$n$}} & \multicolumn{1}{c}{\textbf{$k$}} \\ \midrule
    20 Newsgroup & \numprint[MiB]{575.9} & \numprint{11314} &  \revisedcontent{\numprint{176}} & \numprint{6672} & 20 \\
    Emails & \numprint[MiB]{486.0} & \numprint{9111} & \revisedcontent{\numprint{182}} & \numprint{6992} & 4 \\
    GitHub Projects & \numprint[MiB]{2024.5} & 653 &  \revisedcontent{\numprint{52635}} & \numprint{405117} & 8 \\
    %Ecommerce & \numprint[MB]{1783.9} & \numprint{50123} & \numprint{4665} & 4 \\
    Reuters & \numprint[MiB]{205.5} & \numprint{9122} & \revisedcontent{\numprint{102}} & \numprint{2953} & 65 \\
    Seven Categories & \numprint[MiB]{993.8} & \numprint{3127} & \revisedcontent{\numprint{396}} & \numprint{11373} & 7 \\
    
    \bottomrule
    \end{tabular}
    \label{tab:datasets}
\end{table}

Neither the surveys nor the evaluations of the existing benchmarks consider TMs as components of the layout process.
Even for DRs developed for visualizing text corpora, TMs were not considered in their evaluation~\cite{ingram2015dimensionality,Chen2009InfoVis}.
Our work addresses this gap by evaluating a benchmark that explicitly considers text corpora and TMs as essential layout components.
The idea of such a benchmark was previously proposed by Atzberger and Cech et al.~\cite{atzberger2022-topicmodel-benchmark}.
The authors proposed a benchmark $\mathcal{B} = (\mathcal{D}, \mathcal{L} ,\mathcal{Q})$, consisting of a set of text corpora $\mathcal{D}$, a set of layouts $\mathcal{L}$, and a set of quality metrics $\mathcal{Q}$.
We revised and extended this approach with respect to the following three aspects:
(1) we only consider corpora in $\mathcal{D}$, where the given categories correspond to semantic concepts.
(2) Our layouts $\mathcal{L}$ contain additional TMs but focus on a subset of the proposed DRs.
(3) In addition to accuracy metrics, our set of quality metrics $\mathcal{Q}$ further contains measures for quantifying the perceptual effectiveness of the resulting layout.

\subsection{\texorpdfstring{Datasets $\mathcal{D}$}{Datasets}}

Our set $\mathcal{D}$ contains five corpora, whose characteristics are summarized in \Cref{tab:datasets}.
Each element $D = (C, P) \in \mathcal{D}$ is given as a pair consisting of a corpus $C = \{d_1, \dots, d_m \}$ and a partition $P = \{c_1, \dots, c_k\}$, i.e., a disjoint decomposition of $C$ in $k$ classes.
The corpus $C$ consisting of $m$ documents $d_1, \dots, d_m$ over a vocabulary $\mathcal{V}_{D}$ of size $n = \vert \mathcal{V}_{D} \vert$ is given as a DTM, in which the entry in cell $(i,j)$ indicates the frequency of the term $w_j \in \mathcal{V}_{D}$ in document $d_i$.
The elements in a class $c \in P$ are documents that belong to a higher-level concept.

Four of the five corpora in $\mathcal{D}$ are preprocessed versions of the commonly used datasets \textit{20 Newsgroup}\footnote{\href{https://scikit-learn.org/0.19/datasets/twenty_newsgroups.html}{scikit-learn.org/0.19/datasets/twenty\_newsgroups.html}}, \textit{Emails}\footnote{\href{https://www.kaggle.com/datasets/dipankarsrirag/topic-modelling-on-emails}{kaggle.com/datasets/dipankarsrirag/topic-modelling-on-emails}}, \textit{Reuters}\footnote{\href{https://www.kaggle.com/datasets/nltkdata/reuters}{kaggle.com/datasets/nltkdata/reuters}}, and \textit{Seven Categories}\footnote{\href{https://www.kaggle.com/datasets/deepak711/4-subject-data-text-classification}{kaggle.com/datasets/deepak711/4-subject-data-text-classification}}.
We applied standard preprocessing steps, e.g., removal of stop words and the lemmatization of the vocabulary, and additional, dataset-dependent steps, e.g., removing the email header in the 20 Newsgroup dataset.
For details, we refer to our Git repository.
Our fifth dataset covers the domain of software visualization.
Previous work has shown that source code is suitable for text analysis by TMs~\cite{ascstd2022-codecv,chen2016survey}.
As one corpus on source code, we propose a \textit{GitHub Projects} corpus containing 653 documents from eight categories, i.e., where each document contains the merged source code files of a software project that belongs to one particular \textit{GitHub topic}\footnote{\href{https://github.com/topics}{github.com/topics}}.
\revisedcontent{Based on the file extension, we detect source code files written in one of the following languages: \textit{C}, \textit{C++}, \textit{C\#}, \textit{Go}, \textit{Java}, \textit{JavaScript}, \textit{Move}, \textit{PHP}, \textit{Python}, \textit{Ruby}, \textit{Rust}, or \textit{Solidity}.}
Thereby, we collected the 100 most popular projects ranked by stars\footnote{\href{https://docs.github.com/en/get-started/exploring-projects-on-github/saving-repositories-with-stars}{docs.github.com/en/get-started/exploring-projects-on-github/saving-repositories-with-stars}} for each of the following GitHub topics: \textit{cryptocurrency}, \textit{data-visualization}, \textit{machine-learning}, \textit{frontend}, \textit{database}, \textit{shell}, \textit{server}, and \textit{3d}.
As we assume a disjoint partitioning, we consider only the first mention of a project within the query results.
\revisedcontent{Most remarkable is the large size of the vocabulary $n$ as shown in \Cref{tab:datasets}. This is because, before preprocessing, terms are included, such as short identifier names, which do not occur in English.}
In addition to the usual preprocessing, source code-specific operations are performed, such as separating identifier names according to standard naming conventions and filtering keywords of programming languages as they carry no semantic meaning.
\revisedcontent{Ideally, after preprocessing, the vocabulary contains all English language words that occur as comments or identifiers. However, whether the term originates from a comment or an identifier name is not distinguished. This procedure is typical for the application of topic models to source code~\cite{chen2016survey}.}

\subsection{\texorpdfstring{Layouts $\mathcal{L}$}{Layouts}}
The elements of the set of layouts $\mathcal{L}$ originate from combinations of a TM and a subsequent DR.
Training a TM on a given corpus, given by a DTM, yields a document representation in a Euclidean standard space of dimensionality $\le n$.
The high-dimensional document representations are projected using a DR on a two-dimensional plane.
Particular combinations applied in existing visualization approaches are summarized in \Cref{tab:vis techniques}.

\subsubsection{Topic Models}

From the DTM description of a corpus, each document is given as an $n$-dimensional vector containing the absolute frequencies of each term.
Together with a similarity measure between documents, this representation is denoted as the \textit{Vector Space Model} (VSM)~\cite{crain2012dimensionality}.
In our considerations, we use the cosine similarity for documents, as it allows the comparison of documents of different lengths.
However, the DTM only contains the absolute frequencies of a term regardless of whether the term is also frequently represented in other documents.
In practice, however, the terms represented in only a few documents often indicate an underlying concept and are particularly relevant.
By weighting the entries in the DTM according to the \textit{term frequency-inverse document frequency} (tf-idf) scheme, the VSM can be modified to incorporate this effect~\cite{aggarwal2012surveyClustering}.
Specifically, the tf-idf of a term $w$ in document $d \in C$ is given by the product of the term-frequency of the term $w$ in $d$ and the inverse-document-frequency of $w$ in $d$, i.e.,
\begin{align}
    \text{tf-idf}(w,d)
    = \dfrac{n(w,d)}{\sum \limits_{d' \in C} n(w, d')} \cdot \log \Bigg ( \dfrac{\vert C \vert}{\vert \{d' \in C\vert w \in d'\} \vert} \Bigg),
\end{align}
where $n(w,d)$ denotes the frequency of term $w$ in document $d$.

The DTM is a sparse matrix, i.e., most entries are zero, as documents usually contain a small fraction of the entire vocabulary.
Most TMs aim to find a more compressed representation of the DTM by grouping co-occurring words into topics.
\revisedcontent{Algorithms that detect topics as part of their results are called topic models.
Topics are thereby given as vectors of size $n$, with the $i^{\text{}th}$ entry containing a weight that describes the impact of term $w_i$ for the topic.
From the most relevant words, a human-interpretable concept can be inferred in most cases.
In that sense, the VSM and its tf-idf weighted variant are not a topic model.
In our considerations, we will cover four different topic models described below.
However, in the following, when talking about the evaluation of all TMs, we also include the VSM and its tf-idf weighted variant.}
For example, \textit{Latent Semantic Indexing} (LSI) is based on Singular Value Decomposition (SVD), which results in a decomposition of a given DTM into a document-topic matrix and a topic-term matrix~\cite{deerwester1990indexing}.
In practice, tf-idf weighting is often applied to increase the interpretability of the topics.
Another linear algebra approach for topic modeling is \textit{Non-Negative Matrix Factorization} (NMF), where the DTM or its tf-idf-weighted variant is approximated as a product of two matrices, i.e., a document-topic matrix and a topic-term matrix~\cite{lee1999learning}.
\textit{Latent Dirichlet Allocation} (LDA) is a probabilistic approach for topic modeling and is probably the most widely used TM in the visualization domain.
LDA is based on the assumption of a generative process underlying a corpus.
Training an LDA model results in topics that are given as multinomial distributions over the vocabulary. 
Further, each document is represented as a multinomial distribution over the topics~\cite{blei2003latent}.
As documents are given as distributions, we specifically apply the Jensen-Shannon distance for measuring the similarity between documents.
As LDA is a probabilistic model, we do not replace the DTM with the tf-idf weighted entries.
As the last TM, we integrated \textit{Bidirectional Encoder Representations from Transformers} (BERT), which is a deep learning-based approach for topic modeling that is known to generate easily interpretable topics~\cite{devlin2018bert}.
Unlike the other TMs, each document is described as a high-dimensional vector associated with exactly one or zero topics.
The topics are then derived from these associations using a class-based tf-idf weighting.
In the case of BERT, the similarity between documents is again given by the cosine similarity.
According to our survey of works, these methods are representative of topic extraction in a significant part of the document visualization literature.

\subsubsection{Dimensionality Reductions}
As DRs, we consider t-SNE and UMAP, as they have shown promising results in earlier studies~\cite{Espadoto2021Toward}.
We further consider MDS and SOMs, as they are widely used in the text visualization domain, as shown in \Cref{tab:vis techniques}.
\revisedcontent{Although many more dimension reductions exist, we limit our considerations to these four for capacity reasons.}
\textit{t-distributed Stochastic Neighbor Embedding} (t-SNE) is a DR designed to preserve local structures within a dataset~\cite{van2008visualizing}.
This is accomplished by assuming a Gaussian distribution centered around each point in the given high-dimensional space, representing the probability of picking another point as a neighbor.
The number of effective neighbors considered is controlled by the perplexity hyperparameter, which allows to trade off local and global properties.
The main goal of t-SNE is the preservation of neighborhoods in the low-dimensional representation with respect to a t-distribution.
The final layout is obtained by an iterative optimization process that minimizes a stress function that measures the difference in overall similarity scores derived from the respective distributions. %typically using the Barnes-Hut algorithm.
\textit{Uniform Manifold Approximation and Projection} (UMAP) was developed to address the shortcomings of t-SNE, e.g., the distances between clusters in a t-SNE plot allow no interpretation~\cite{mcinnes2020umap}.
Conceptually similar to t-SNE, UMAP differs in its mathematical details, e.g., it relies on a stress function derived from Cross-Entropy rather than the Kullback-Leibler divergence.
UMAP has two hyperparameters: the number of neighbors as a trade-off between preserving local and global structures and the minimal distance that controls how close data points can be grouped together in the two-dimensional layout.
\textit{(Metric) Multidimensional Scaling} (MDS) operates on a dissimilarity matrix of the dataset, i.e., a matrix that contains the pairwise distances between the data points.
MDS aims to compute a lower-dimensional representation, such that the pairwise Euclidean distances between the points in the layout reflect the entries in the dissimilarity matrix~\cite{cox2008multidimensional}.
In particular, MDS allows for the visualization of abstract datasets that are not embedded in the Euclidean space.
The positions of the data points are computed iteratively by optimizing a stress function, for example, using the SMACOF algorithm.
The number of iterations is the only hyperparameter of the model.
\textit{Self-Organizing Maps} (SOMs) are a class of fully-connected two-layered neural networks where the neurons of the second layer are arranged on a two-dimensional grid, whose width and height are given by two hyperparameters~\cite{kohonen1997som}.
For a given input, the neuron whose weight vector is most similar to the input is activated.
This so-called best matching unit determines the position of the given input vector in the two-dimensional space.
The weights are adjusted during the training phase starting from random initialization in order to minimize the sum of all quantization errors, i.e., the differences between the input vectors and their best matching unit.
In the case of the SOM, we applied a PCA that captures \numprint[\%]{95} of the variance for a given dataset to reduce computational efforts~\cite{jolliffe2005principal}.

The combination of a TM with a DR allows for special considerations.
For example, in the case of LDA, the similarities between the topics differ, as they are given by multinomial distributions over the vocabulary.
Applying a DR on the document-topic representation does not consider those similarities, thus treating the topics as orthogonal to each other.
Atzberger et al.\ proposed an alternative by first applying the DR on the topics and then aggregating the positions of the documents as linear combinations according to their document-topic representation~\cite{atzberger2021softwareforest}.
The position $\bar{d}$ of document $d$ is therefore given by
\begin{align} \label{eq: convex combination}
    \bar{d} = \sum \limits_{j=1}^{K} \theta_j \bar{\phi}_j,
\end{align}
where $\theta = (\theta_1, \dots, \theta_K)$ denotes the topic representation of $d$, and $\bar{\phi}_1, \dots, \bar{\phi}_K$ denotes the positions of the topics after application of a DR.

\subsection{\texorpdfstring{Quality Metrics $\mathcal{Q}$}{Quality Metrics}}
The elements in the set $\mathcal{Q}$ are quality measures that quantify certain layout aspects.
Similar to previous benchmark studies, we measure the quality of a layout with respect to the local and global structures of a corpus by using several accuracy metrics.
The \revisedcontent{\textit{Trustworthiness}} $\alpha_{T}$ measures the percentage of close points in the two-dimensional layout that are also close in the VSM~\cite{venna2006visualizing}.
Vice versa, the \textit{Continuity} $\alpha_{C}$ measures the percentage of points in the VSM that are also close in the two-dimensional layout~\cite{venna2006visualizing}.
For both metrics, we refer to the seven nearest neighbors, as suggested in previous studies~\cite{Espadoto2021Toward,vernier2020quantitative,vernier2021guided}.
The \textit{7-Neighborhood hit} $\alpha_{NH}$ requires labels for each document.
It measures the percentage of points with the same label among the seven nearest neighbors, averaged over all points~\cite{Paulovich2006PBC}.
All three metrics have values in the $[0,1]$ range, with 1 being the optimal score.
The last accuracy metric is based on the \textit{Shephard Diagram}, a two-dimensional scatter plot that relates the pairwise distances in $D$ to the Euclidean distances in the layout~\cite{Joia2011LocalAffine}.
In an ideal scenario, the Shephard Diagram would be a subset of the diagonal.
Then, the \textit{Shephard Digram Correlation} $\alpha_{SDC}$ is a measure for the deviation from the ideal scenario.
It is given by the Spearman Rank Correlation of the Shephard Diagram.
The metric ranges between $[-1,1]$, with 1 being the optimal score.

As a second group of metrics, we approximate the effectiveness of perceiving resulting clusters and existing categories.
Following the results of Sedlmair and Aupetit, we include the \textit{Distance Consistency} $\beta_{DC}$~\cite{sedlmair2015data}.
It measures the percentage of points \revisedcontent{in the projected two-dimensional space} whose category center, i.e., the average of all points in that category, is also its nearest category center~\cite{sips2009selecting}.
The \textit{Silhouette Coefficient} $\beta_{SC}$ compares the mean intra-cluster distance $a$ and the mean inter-cluster distance $b$, where in our case, the cluster labels are given by the categories~\cite{Rousseeuw1987Silhouettes}.
By dividing the difference $b-a$ by $\max(a,b)$, the Silhouette Coefficient ranges between $[-1,1]$, with 1 being the optimal score.
We further apply the \textit{Calinski-Harabasz index} $\beta_{CH}$~\cite{Calinski1974} and the \textit{Davies-Bouldin-index} $\textit{DB}$~\cite{Davies1979}.
In both cases, the metric results in a non-negative value.
In our benchmark, we normalize both by dividing them by the maximal value achieved on a dataset-TM combination, with 1 being the optimal score for $\beta_{CH}$ and 0 being the optimal score for $\beta_{DB}$.

\subsection{Experimental Setup \& Implementation}
\label{section: Implementation}
We implemented this benchmark using Python 3 and state-of-the-art libraries for all TMs and DRs.
Regarding deployment, the implementation was designed for concurrent execution on a computational cluster.
The source code of the benchmark, including the scripts to generate the GitHub Projects dataset, is available in a Git repository\footnotemark[1].

\subsubsection{Hyperparameter Settings}

\begin{table}[t]
    %\tiny
    \footnotesize
    \centering
	\setlength{\tabcolsep}{4.0pt}%
	\renewcommand{\arraystretch}{1.00}%
    \caption{Range for the hyperparameters considered in our experiments. Each configuration for one DR is combined with a dataset and TM.}
    \vspace{-0.5\baselineskip}
    \begin{tabular}{rrr}
    \toprule
    \multicolumn{1}{c}{\textbf{DR}} & \multicolumn{1}{c}{\textbf{Parameter Name}} & \multicolumn{1}{c}{\textbf{Values}} \\ \midrule
    t-SNE & \texttt{learning\_rate} & 250, 1000, 2000, 4000, 10000 \\
    t-SNE & \texttt{n\_iter} & 10, 17, 28, 46, 77, 129, 215, 359, 599, 1000 \\
    t-SNE & \texttt{perplexity} & 5--50 step size 5 \\[0.6ex] %\midrule
    UMAP & \texttt{min\_dist} & 0.0--1.0 step size 0.1 \\
    UMAP & \texttt{n\_neighbors} & 2, 10, 11, 12, 13, 14, 15, 16, 17, 18, 19, 20 \\[0.6ex] %\midrule
    SOM & \texttt{m} & 10--20 step size 1 \\
    SOM & \texttt{n} & 10--20 step size 1 \\[0.6ex] %\midrule
    MDS & \texttt{max\_iter} & 300--900 step size 20 \\
    
    \bottomrule
    \end{tabular}
    \label{tab:parameters}
\end{table}

For each dataset, we consider precisely one version of each TM, using a fixed hyperparameter configuration chosen using best practices~\cite{Wallach2009RethinkingMatter}.
In the case of just a few categories $k$, i.e., for the Emails corpus, the Seven Categories corpus, and the GitHub Projects corpus, we set the number of topics $K=2k$.
For the 20 Newsgroup corpus and the Reuters corpus, we set $K = k$.
As a general check for plausibility, we further inspected the most relevant words for each topic with respect to interpretability.
We limited the benchmark regarding the TMs as iterating over the hyperparameters of each TM, e.g., the number of topics, or the Dirichlet priors of the LDA model, would enlarge the benchmark by multiple orders of magnitude.
The ranges for the hyperparameters of the four different DRs are summarized in \Cref{tab:parameters}.
In principle, the validation of the individual hyperparameters follows a grid search, resulting in above \numprint{45000} different hyperparameter configurations.
However, the order of the layout computation is random to ensure that representative results are available during preliminary analysis.

\subsubsection{Software Dependencies}
Our implementation is based on Python 3.10 and several actively maintained and widely used third-party libraries.
For LDA, LSI, and NMF, we have chosen the \textit{Gensim} library (4.2.0), and for BERT, the \textit{Sentence Transformer} library (2.2.2).
\revisedcontent{As the application of BERT relies on pretrained word embeddings, it is a deterministic approach to topic modeling. The algorithms provided by the Gensim implementation are also deterministic after initialization.}
We have chosen the \textit{distilbert-base-nli-mean-tokens} as Sentence Transformer\footnote{\href{https://metatext.io/models/sentence-transformers-distilbert-base-nli-stsb-mean-tokens}{metatext.io/models/sentence-transformers-distilbert-base-nli-stsb-mean-tokens}}.
For t-SNE and MDS, we use \textit{Scikit-Learn} (1.2.1) for the Silhouette Coefficient, the Calinski-Harabasz index, and the Davies-Bouldin index as well.
For UMAP, we use the \textit{UMAP-Learn} library (0.5.3).
For the SOM, we have chosen the implementation provided by \textit{Sparse-SOM} (0.6.1)~\cite{Melka2019}.
Furthermore, our preprocessing is based on the libraries \textit{NLTK} (3.7), e.g., for removal of stop words, and \textit{Spacy} (3.4.3) for lemmatization.

\subsubsection{Computational Cluster}
We set up the benchmark on a computational cluster with \textit{Simple Linux Utility for Resource Management} (SLURM) \cite{SLURM} for concurrent execution.
This cluster allowed for a large speed-up while requiring special handling for the software deployment and job scheduling.
We had access to the AMD~x64 nodes, which came in two kinds of hardware configurations:
(1) HPE XL225n~Gen10 machines with 2 AMD EPYC~7742 processors, \numprint[GiB]{512} RAM, and 64 cores, and
(2) Fujitsu RX2530~M5 machines with 2 Intel Xeon~Gold~5220S processors, \numprint[GiB]{96} RAM, and 32 cores.
From the HPE XL225n~Gen10 nodes, we regularly used 11, and from the Fujitsu RX2530~M5, we regularly used 10.
The SLURM setup on each node required to use an Enroot\footnote{\href{https://github.com/NVIDIA/enroot}{\faGithub{} NVIDIA/enroot}} container with the benchmark.
We derived this Enroot container from a Docker\footnote{\href{https://www.docker.com/}{docker.com}} container using Pyxis\footnote{\href{https://github.com/NVIDIA/pyxis}{\faGithub{} NVIDIA/pyxis}} and reused the container via caching.
\par
Scheduling a job via SLURM requires explicit specification of required resources, e.g., RAM, as upper limits.
Given our heterogeneous dataset sizes (\Cref{tab:datasets}), we decided to split up job allocations into two running sets, a memory-heavy and a memory-moderate one.
The memory-heavy set covers the evaluation for the GitHub Projects dataset with a RAM allocation of \numprint[GiB]{200}.
The memory-moderate set covers the other datasets with a RAM allocation of \numprint[GiB]{40}.
Overall, using this design, we ran thousands of jobs corresponding to ten thousands of different layouts on the cluster.
On average, a job of the memory-heavy set had an execution time of \numprint[hours]{12} with a maximum of \numprint[hours]{36}.
For the memory-moderate set, the average execution time was around \numprint[hours]{2}, with a maximum of \numprint[hours]{10}.

\section{Results}
\label{section: results}
We evaluated \numprint{46311} samples, which corresponds to $\approx$\numprint[\%]{94.7} of targeted layouts.
The remaining \numprint[\%]{5.3} of targeted layouts could not be computed due to exceeding memory consumption.
The corresponding quality metrics are stored as a tabular dataset.
Further, the dataset is augmented with two additional aggregated quality metrics concerning accuracy and perception.
The aggregated quality metrics are given as linear combinations of the accuracy and perception metrics, respectively, taking into account a correlation analysis to limit the influence of strongly correlated quality metrics. %  (see below)
We then performed an analysis of this dataset with respect to \revisedcontent{four} specific questions: %  (RQ)
\begin{enumerate}
    \setlength{\itemsep}{0pt}
    \item \revisedcontent{Do the tf-idf weighting scheme and \Cref{eq: convex combination} improve the results?}
    \item Which layout achieves the best result for a given dataset with respect to accuracy or perception?
    \item How sensitive are the DRs with respect to their hyperparameters?
    \item What is the performance of the default hyperparameters?
\end{enumerate}

\subsection{Correlation of Quality Metrics}

\begin{figure}[t]
    \vspace{-5pt}
    \pgfplotstableread[col sep=comma]{data/Results_Correlation_Final_matrix.csv}\loadeddata%
    \setlength{\tabcolsep}{3pt}%
    \renewcommand{\arraystretch}{1.3}%
    \pgfplotstabletypeset[%
        font=\footnotesize,
        col sep=&,
        row sep=\\,
        columns={
            metric,
            Trustworthiness,
            Continuity,
            Shephard Diagram Correlation,
            7-Neighborhood Hit,
            Calinski-Harabasz-Index Normalized,
            Davies-Bouldin-Index Normalized,
            Distance Consistency,
            Silhouette Coefficient,
            ColorBar
        },
        columns/metric/.style={
            string type,
            column type={r},
            column name={\strut},
            string replace={Trustworthiness}{\revisedcontent{\strut$\alpha_{T}$}},
            string replace={Continuity}{\revisedcontent{\strut$\alpha_{C}$}},
            string replace={Shephard Diagram Correlation}{\revisedcontent{\strut$\alpha_{SDC}$}},
            string replace={7-Neighborhood Hit}{\revisedcontent{\strut$\alpha_{NH}$}},
            string replace={Distance Consistency}{\revisedcontent{\strut$\beta_{DC}$}},
            string replace={Calinski-Harabasz-Index Normalized}{\revisedcontent{\strut$\beta_{CH}$}},
            string replace={Davies-Bouldin-Index Normalized}{\revisedcontent{\strut$\beta_{DB}$}},
            string replace={Silhouette Coefficient}{\revisedcontent{\strut$\beta_{SC}$}},
        },
        columns/Trustworthiness/.style={
            colorbarstyle,
            column name={ \revisedcontent{$\alpha_{T}$} },
            color cells={min=-1,max=1},
            /pgfplots/colormap name=RdBu,
        },
        columns/Continuity/.style={
            colorbarstyle,
            column name={ \revisedcontent{$\alpha_{C}$} },
            color cells={min=-1,max=1},
            /pgfplots/colormap name=RdBu,
        },
        columns/Shephard Diagram Correlation/.style={
            colorbarstyle,
            column name={ \revisedcontent{$\alpha_{SDC}$} },
            color cells={min=-1,max=1},
            /pgfplots/colormap name=RdBu,
        },
        columns/7-Neighborhood Hit/.style={
            colorbarstyle,
            column type/.add={}{@{\hspace{11pt}}},
            column name={ \revisedcontent{$\alpha_{NH}$} },
            color cells={min=-1,max=1},
            /pgfplots/colormap name=RdBu,
        },
        columns/Distance Consistency/.style={
            colorbarstyle,
            column name={ \revisedcontent{$\beta_{DC}$} },
            color cells={min=-1,max=1},
            /pgfplots/colormap name=RdBu,
        },
        columns/Calinski-Harabasz-Index Normalized/.style={
            colorbarstyle,
            column name={ \revisedcontent{$\beta_{CH}$} },
            color cells={min=-1,max=1},
            /pgfplots/colormap name=RdBu,
        },
        columns/Davies-Bouldin-Index Normalized/.style={
            colorbarstyle,
            column name={ \revisedcontent{$\beta_{DB}$} },
            color cells={min=-1,max=1},
            /pgfplots/colormap name=RdBu,
        },
        columns/Silhouette Coefficient/.style={
            colorbarstyle,
            column name={ \revisedcontent{$\beta_{SC}$} },
            color cells={min=-1,max=1},
            /pgfplots/colormap name=RdBu,
        },
        multicolumn names=c,
        %every head row/.style={
        %    before row={\toprule},
        %    after row=\midrule
        %},
        %every last row/.style={after row=\bottomrule},
        every row no 3/.style = { before row={\vspace{5pt}} },
        create on use/ColorBar/.style={create col/set={}},
        columns/ColorBar/.style={
            column name={},
            assign cell content/.code={%
            \ifnum\pgfplotstablerow=0%
                \pgfkeyssetvalue{/pgfplots/table/@cell content}%
                {\multirow{\pgfplotstablerows}{*}{
                    \hspace*{0cm}
                    \begin{tikzpicture}[inner sep=0pt,outer sep=0pt]
                        \begin{axis}[
                            hide axis,
                            scale only axis,
                            height=0cm,
                            width=0cm,
                            colormap/RdBu,
                            colorbar horizontal,
                            point meta min=-1,
                            point meta max=1,
                            colorbar style={
                                width=3.4cm,
                                rotate=90,
                                xtick style={draw=gray},
                                xtick={-1,-0.5,0,0.5,1},
                                xticklabels={$-1.0$,$-0.5$,\phantom{$-$}$0.0$,\phantom{$-$}$0.5$,\phantom{$-$}$1.0$},
                                xticklabel style={
                                    font=\footnotesize,
                                    xshift=3pt,
                                    yshift=-0.2ex,
                                    anchor=west,
                                },
                                typeset ticklabels with strut,
                                at={(0.0,0.0)},anchor=south,
                            }]
                            \addplot [draw=none] coordinates {(0,0) (1,1)};
                        \end{axis}
                    \end{tikzpicture}
                }}%
            \else
                \pgfkeyssetvalue{/pgfplots/table/@cell content}{}%
            \fi%
            },
        }
    ]{\loadeddata}
    \vspace{-0.5\baselineskip}
    \caption{Heatmap showing the pairwise correlations between the eight quality metrics using a diverging color scheme.}%
    \label{tab:Heatmap_Correlation}%
\end{figure}
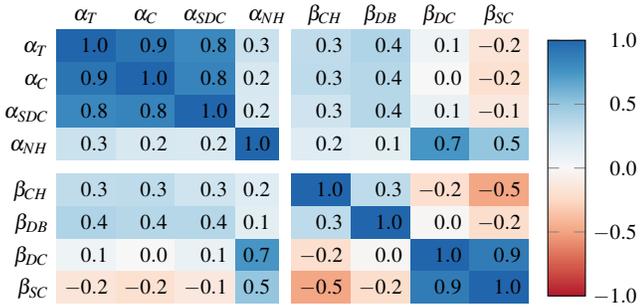

Previous ad-hoc formulations of aggregated metrics relied on the arithmetic mean of the individual metrics, i.e., each metric is weighted equally~\cite{Espadoto2021Toward,vernier2020quantitative,vernier2021guided,Morariu2023PredictingUserPreferences}.
Instead, we target a correlation-adjusted weighting that groups strongly correlated quality metrics using a threshold of \numprint{0.8}.
Specific to the correlation analysis, the David-Bouldin index $\beta_{DB}$ has been replaced by $1-\beta_{DB}$ to achieve its optimal score at 1.
The pairwise correlations are shown in \Cref{tab:Heatmap_Correlation}.
For the accuracy metrics, we observe that the three metrics $\alpha_{T}$, $\alpha_{C}$, and $\alpha_{SDC}$ are strongly positively correlated.
We, therefore, merge them into a single quality metric using their average.
Conversely, the three metrics correlate weakly with $\alpha_{NH}$.
Overall, we define the aggregated accuracy metric $\alpha$ as
\begin{align}
    \alpha = \dfrac{1}{2} \alpha_{NH} + \dfrac{1}{2} \Bigg ( \dfrac{\alpha_{T} + \alpha_{C} + 0.5 \cdot (\alpha_{SDC} + 1)}{3} \Bigg ),
    %= \dfrac{1}{2} \alpha_{NH} + \dfrac{1}{6} \alpha_{T} + 
    %\dfrac{1}{6} \alpha_{C} + \dfrac{1}{6} \alpha_{SDC}
\end{align}
where we replace the Shephard Diagram Correlation $\alpha_{SDC}$ by $0.5 \cdot (\alpha_{SDC} + 1)$ such that it ranges between 0 and 1. 
Our metric $\alpha$ is in the value range $[0,1]$, with 1 being the optimal score.
Regarding the perception metrics, the only strong correlation occurred between $\beta_{DC}$ and $\beta_{SC}$.
The other pairwise correlations do not allow for further grouping.
The aggregated perception metric $\beta$ is defined as
\begin{align}
    \beta = \dfrac{1}{3}(1-\beta_{DB}) + \dfrac{1}{3}\beta_{CH} + \dfrac{1}{3} \Bigg ( \dfrac{0.5 \cdot (\beta_{SC} + 1)+ \beta_{DC}}{2} \Bigg ).
\end{align}
As before, the single metrics are modified such that the aggregated metric $\beta$ ranges between $[0,1]$ with 1 being the optimal score.

\revisedcontent{In our benchmark, we selected datasets with predefined categories.
As the predefined labels indicate a \enquote{higher-level} concept, i.e., semantic themes within the documents, we assume that such a higher-level concept shows relations to the vocabulary, and therefore a TM would yield topics that can be associated with the predefined categories.
A good layout algorithm with respect to the accuracy metric $\alpha$ would result in a two-dimensional scatter plot so that documents that share the same "dominant" topic form a cluster.
Concerning the perception metric, these clusters should be separated well.
Our scatter plots of the best results support this conjecture.}

\subsection{\revisedcontent{Binary Decisions}}

\revisedcontent{The choice of weighting the VSM according to the tf-idf scheme is binary. Using a binary test\footnote{\href{https://docs.scipy.org/doc/scipy/reference/generated/scipy.stats.binomtest.html}{docs.scipy.org/doc/scipy/reference/generated/scipy.stats.binomtest.html}}, we investigate whether the tf-idf weighting improves the accuracy metric $\alpha$ and the perception metric $\beta$.
We extract pairs of layouts from our result dataset where the tf-idf scheme has been applied in one case but not the other, while all the other hyperparameters are the same.
The number of pairs in which the tf-idf improves the results is denoted as k.
The p-values and the lower bounds for the confidence intervals (the upper bound is always one since our unknown parameter is the probability that the tf-idf weighting improves the result) for the confidence level of 0.99 are shown in \Cref{tab:tf-idf_test}.
The p-values of the complete set of pairs, i.e., named Total in \Cref{tab:tf-idf_test}, show that the layout algorithms significantly improve the results concerning $\alpha$ and $\beta$.
However, different results might occur on selected datasets, e.g., in the case of the 7 Categories dataset concerning $\beta$.}

\revisedcontent{Also, the application of \Cref{eq: convex combination} is a binary choice.
Analogously to before, we applied a binary test.
The results are shown in \Cref{tab:lin-comb_test}.
From the p-values of the complete set of pairs, i.e., the total case in \Cref{tab:lin-comb_test}, we conclude that \Cref{eq: convex combination} improves the results concerning accuracy and perception.}

\begin{table}[t]
    %\tiny
    \footnotesize
    \centering
	\setlength{\tabcolsep}{3.5pt}%
	\renewcommand{\arraystretch}{1.00}
    \caption{\revisedcontent{Results of the binary test for the null hypothesis \enquote{The tf-idf weighting scheme does not improve the results according $\alpha$, or $\beta$ respectively.} As the p-values (Total) are 0.00, we reject the zero hypotheses.}}
    \vspace{-0.5\baselineskip}
    \revisedcontent{
    \begin{tabular}{rc|ccc|ccc}
    \toprule
    \multicolumn{2}{c|}{} & \multicolumn{3}{c|}{\textbf{Accuracy Metric $\alpha$}} &  \multicolumn{3}{c}{\textbf{Perception Metric $\beta$}}\\ %\midrule
    \textbf{Dataset} & \textbf{$n$} &
    \textbf{k} & \textbf{p-value} & \textbf{Conf.} &
    \textbf{k} & \textbf{p-value} & \textbf{Conf.}\\ \midrule
    20 Newsgroup & 2842 & 2577 & 0.00 & 0.89 & 2839 & 0.00 & 1.00 \\
    7 Categories & 2897 & 1894 & 0.00 & 0.63 & 655 & 1.00 & 0.21 \\
    Emails & 2909 & 2667 & 0.00 & 0.90 & 2516 & 0.00 & 0.85 \\
    GitHub & 2629 & 2344 & 0.00 & 0.88 & 2370 & 0.00 & 0.89 \\ 
    Reuters & 2895 & 1454 & 0.41 & 0.48 & 1450 & 0.47 & 0.48 \\
    \midrule
    \textbf{Total} & \textbf{14172} & \textbf{10936} & \textbf{0.00} & \textbf{0.76} & \textbf{9830} & \textbf{0.00} & \textbf{0.68}\\
    \bottomrule
    \end{tabular}
    \label{tab:tf-idf_test}
	}
\end{table}

\begin{table}[t]
    %\tiny
    \footnotesize
    \centering
	\setlength{\tabcolsep}{3.5pt}%
	\renewcommand{\arraystretch}{1.00}
    \caption{\revisedcontent{Results of the binary test for the null hypothesis \enquote{\Cref{eq: convex combination} does not improve the results according $\alpha$, or $\beta$ respectively.} As the p-values (Total) are 0.00, we reject the zero hypotheses.}}
    \vspace{-0.5\baselineskip}
    \revisedcontent{
    \begin{tabular}{rc|ccc|ccc}
    \toprule
    \multicolumn{2}{c|}{} & \multicolumn{3}{c|}{\textbf{Accuracy Metric $\alpha$}} &  \multicolumn{3}{c}{\textbf{Perception Metric $\beta$}}\\ %\midrule
    \textbf{Dataset} & \textbf{$n$} &
    \textbf{k} & \textbf{p-value} & \textbf{Conf.} &
    \textbf{k} & \textbf{p-value} & \textbf{Conf.}\\ \midrule
    20 Newsgroup & 3594 & 2404 & 0.00 & 0.65 & 1734 & 0.98 & 0.46 \\
    7 Categories & 3651 & 2117 & 0.00 & 0.56 & 2057 & 0.00 & 0.54 \\
    Emails & 3660 & 2122 & 0.00 & 0.56 & 1753 & 0.99 & 0.46 \\
    GitHub & 3124 & 1898 & 0.00 & 0.59 & 1745 & 0.00 & 0.54 \\ 
    Reuters & 3646 & 2532 & 0.00 & 0.68 & 1841 & 0.28 & 0.49 \\
    \midrule
    \textbf{Total} & \textbf{17675} & \textbf{11073} & \textbf{0.00} & \textbf{0.62} & \textbf{9130} & \textbf{0.00} & \textbf{0.51}\\
    \bottomrule
    \end{tabular}
    }
    \label{tab:lin-comb_test}
\end{table}

\subsection{Optimal Results}
\label{subsec:Optimal_Results}

\begin{table}[t]
    %\tiny
    \footnotesize
    \centering
	\setlength{\tabcolsep}{10.0pt}%
	\renewcommand{\arraystretch}{1.00}%
    \caption{Layout algorithms resulting in the best result with respect to the accuracy metric $\alpha$ for each dataset.}
    \vspace{-0.5\baselineskip}
    \begin{tabularx}{1.0\columnwidth}{rXr}
    \toprule
    \multicolumn{1}{c}{\textbf{Dataset}} & \multicolumn{1}{c}{\textbf{Layout}} & \multicolumn{1}{c}{\textbf{$\alpha$}} \\[0.2ex] \midrule
    20 Newsgroups & (LSI,+,t-SNE,-), (LSI,+,t-SNE,+) & 0.79 \\[0.2ex]
    Emails & (LSI,+,t-SNE,-), (LSI,+,t-SNE,+) & 0.76 \\[0.2ex]
    GitHub Projects & (LSI,+,t-SNE,-),(LSI,+,t-SNE,+) & 0.75\\[0.2ex]
    Reuters & (LDA,X,t-SNE,+) & 0.66 \\[0.2ex]
    Seven Categories & \makecell[lt]{(LSI,-,t-SNE,-), (LSI,-,t-SNE,+) \\ (LSI,+,t-SNE,-), (LSI,+,t-SNE,+) \\ (LSI,-,UMAP,-), (LSI,+,UMAP,-)} & 0.78 \\
    \bottomrule
    \end{tabularx}
    \label{tab:optimal alpha values}
\end{table}

\begin{table}[t]
    %\tiny
    \footnotesize
    \centering
	\setlength{\tabcolsep}{10.0pt}%
	\renewcommand{\arraystretch}{1.00}%
    \caption{Layout algorithms resulting in the best result with respect to the perception metric $\beta$ for each dataset.}
    \vspace{-0.5\baselineskip}
    \begin{tabularx}{1.0\columnwidth}{rXr}
    \toprule
    \multicolumn{1}{c}{\textbf{Dataset}} & \multicolumn{1}{c}{\textbf{Layout}} & \multicolumn{1}{c}{\textbf{$\beta$}} \\ \midrule
    20 Newsgroups & (VSM,+,t-SNE,X), (LSI,+,t-SNE,+) & 0.80 \\[0.2ex]
    Emails & (VSM,+,t-SNE,X) & 0.87 \\[0.2ex]
    GitHub Projects & (VSM,-,t-SNE,X) & 0.77\\[0.2ex]
    Reuters & (VSM,+,t-SNE,X) & 0.69 \\[0.2ex]
    Seven Categories & (VSM,+,t-SNE,X) & 0.80 \\
    \bottomrule
    \end{tabularx}
    \label{tab:beta values}
\end{table}

\begin{figure}[tp]
    \centering
    \includegraphics[width = 0.99\linewidth,trim={3.5cm 0.7cm 3.5cm 0.23cm},clip]{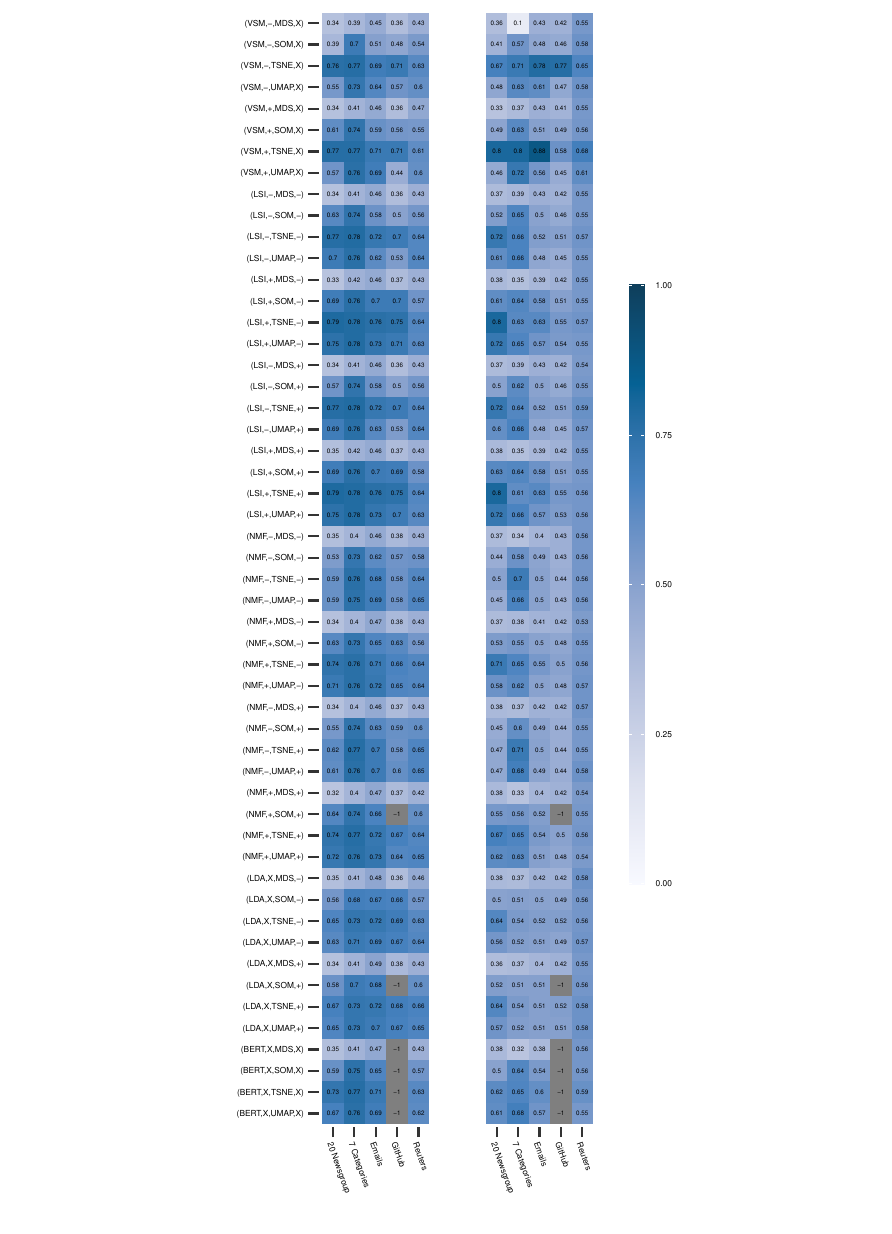}
    \caption{Heatmap showing the best results for the layout algorithms on each dataset. The grey cells indicate combinations, where the TM could not be applied on the dataset. \revisedcontent{The second entry of quadruple specifying the layout indicates, whether the tf-idf weighting was applied (+), or not (-), or could not be applied (X). The fourth entry in the quadruple indicates whether \Cref{eq: convex combination} was applied (+), or not (-), or could not be applied (X).} (Left) Optimal values for the accuracy metric $\alpha$, (Right) Optimal values for the perception metric $\beta$.}
    \label{fig:heatmap}
\end{figure}

We first consider the optimal values for $\alpha$ and $\beta$ for each layout on each dataset.
The results are summarized in  \Cref{fig:heatmap}.
Some layouts could not be computed (grey cells) due to exceeding memory consumption.
For example, BERT is based on a pre-trained embedding, which has not been shown suitable for modeling source code.
This is because the use of identifiers in source code differs from natural language as identifiers without vocals might have a semantic meaning, e.g., \enquote{ccxt} has no vocal and is, therefore, no \enquote{natural} word but refers to a cryptocurrency trading API that is well known among practitioners.

We assigned each layout algorithm an identifier given as a quadruple.
The first entry indicates the TM.
The second entry indicates whether the tf-idf weighting was applied (+), or not (-), or could not be considered for the specific TM (X).
The third entry contains the DR.
The fourth entry indicates whether the position of the documents was computed according to \Cref{eq: convex combination} as a linear combination (+), or not (-), or could not be considered, as no topics were extracted (X).

\subsubsection{What Layouts perform best for a given dataset?}
% Accuracy
The optimal results for each dataset are summarized in \Cref{tab:optimal alpha values}.
No layout algorithm scores best over all datasets with respect to $\alpha$.
In any case, the first two entries of the quadruple are either given by (LSI,+), (LSI,-), or (LDA, X).
We suspect the reduction in dimensionality within LSI and LDA to be an advantage for a subsequent DR compared to the VSM and BERT.
The last two entries specifying the best DR are either given by (t-SNE,+), (t-SNE,-), or (UMAP,+).
This observation is aligned with the results of Espadoto et al. in the case of the VSM~\cite{Espadoto2021Toward}.
However, most of the layout algorithms perform similarly.
Only MDS shows a generally lower performance compared to the other DRs.
The optimal results with respect to $\beta$ are summarized in \Cref{tab:beta values}.
As before, the best results are achieved when t-SNE is applied.
However, contrary to our observations on $\alpha$, LDA and LSI perform \revisedcontent{inferior to} the baseline VSM approach.
%In the case of any other DR, applying a TM can improve the overall quality by up to XXX percent.
As our metric $\beta$ mainly quantifies the cluster separation, this observation confirms the prevailing opinion that t-SNE produces well-separated clusters.

\subsubsection{What is the influence of the parameters $n$, $m$, $k$?}
Regarding $\alpha$, in \revisedcontent{\numprint[\%]{71.2}} of the cases, the best result for a given layout is achieved on the Seven Categories corpus, which is characterized by a small number of documents $m$.
In \revisedcontent{\numprint[\%]{28.8}} of the cases, the layout does not perform best on the Seven Categories corpus.
In that case, \revisedcontent{\numprint[\%]{86.7}} use MDS as DR.
Although the GitHub Projects corpus contains fewer documents, we suspect its high dimensionality $n$ is why the best result for a layout is never achieved on the GitHub Projects corpus.
With respect to $\beta$ in \revisedcontent{\numprint[\%]{44.3}} of the rows, the Seven Categories corpus achieves the optimal result, followed by the Reuters (\revisedcontent{\numprint[\%]{36.5}}) and 20 Newsgroup corpora (\revisedcontent{\numprint[\%]{19.2}}), and the Emails (\revisedcontent{\numprint[\%]{3.8}}).
In no case, the optimal result is achieved on the GitHub Projects dataset.
This order does not correspond to the increasing number of categories $k$.
We assume Seven Categories performs best because of its small number of documents.
We suspect that in these cases, the vocabulary has a more direct relationship to the topics, and thus more distinct clusters emerge when modeling the corpora using TMs.

\subsubsection{Which DR technique performs best for a given TM?}
Regarding $\alpha$, for any of the given TMs, the best performing DR is either UMAP in \revisedcontent{\numprint[\%]{27.0}} of the cases or t-SNE in \revisedcontent{\numprint[\%]{81.0}} of the cases.
A similar trend is obtained with respect to $\beta$.
Here, t-SNE achieves the optimal result in \revisedcontent{\numprint[\%]{84.1}} of the cases and UMAP in \revisedcontent{\numprint[\%]{23.8}} of the cases.
However, in the cases where UMAP is superior to t-SNE, their difference is negligible.
With respect to $\alpha$, in \revisedcontent{\numprint[\%]{47.5}} of the cases, the layout algorithms applying a linear combination according to \Cref{eq: convex combination} perform equally to the direct application of the DR on the document representation.
In \revisedcontent{\numprint[\%]{36.6}} of cases, using the linear combination improves the result, and in \revisedcontent{\numprint[\%]{15.9}} of cases, it performs less.
Similar results are observed with respect to $\beta$.
In \revisedcontent{\numprint[\%]{31.2}} of the cases, it improves the results.
In \revisedcontent{\numprint[\%]{44.5}} of the cases, it matches the results and performs less in the remaining \revisedcontent{\numprint[\%]{24.3}} of the cases.

\subsection{Influence of the Hyperparameters}
\begin{figure}[t]
    \centering
    \includegraphics[width = 0.9\linewidth]{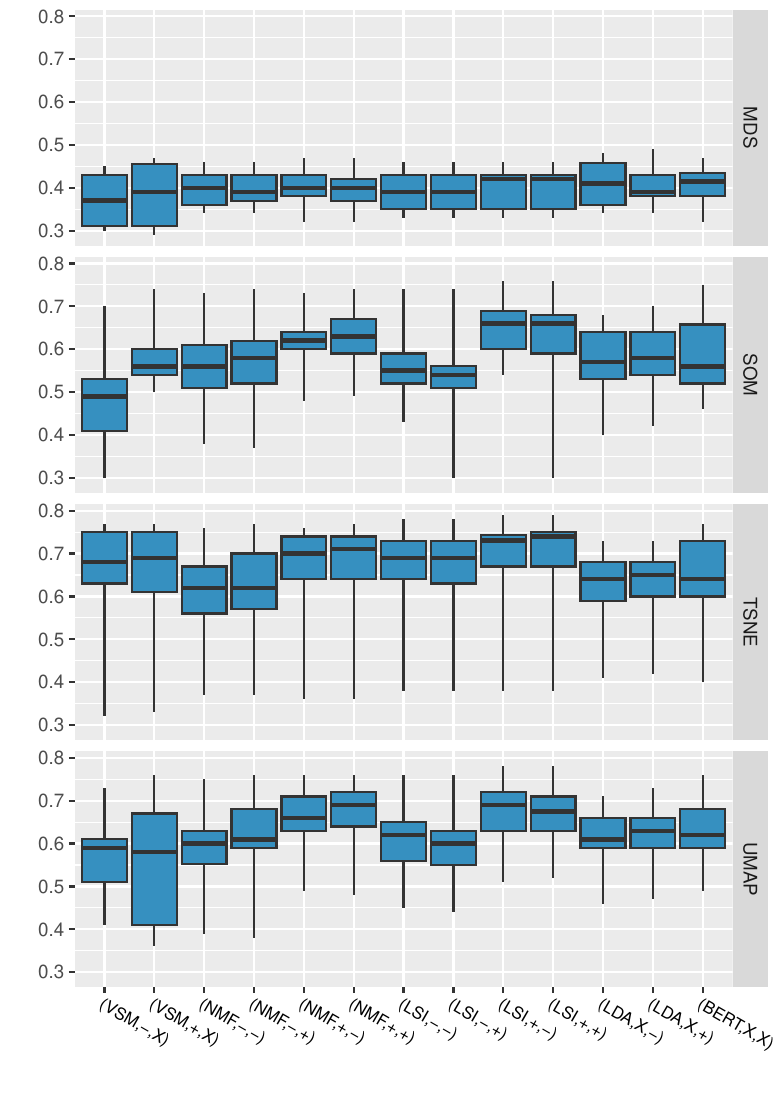}
    \vspace{-1.5\baselineskip}
    \caption{Boxplots showing the summary statistic of the quality metric $\alpha$.}
    \label{fig:Boxplot_Accuracy}
\end{figure}

\begin{figure}[t]
    \centering
    \includegraphics[width = 0.9\linewidth]{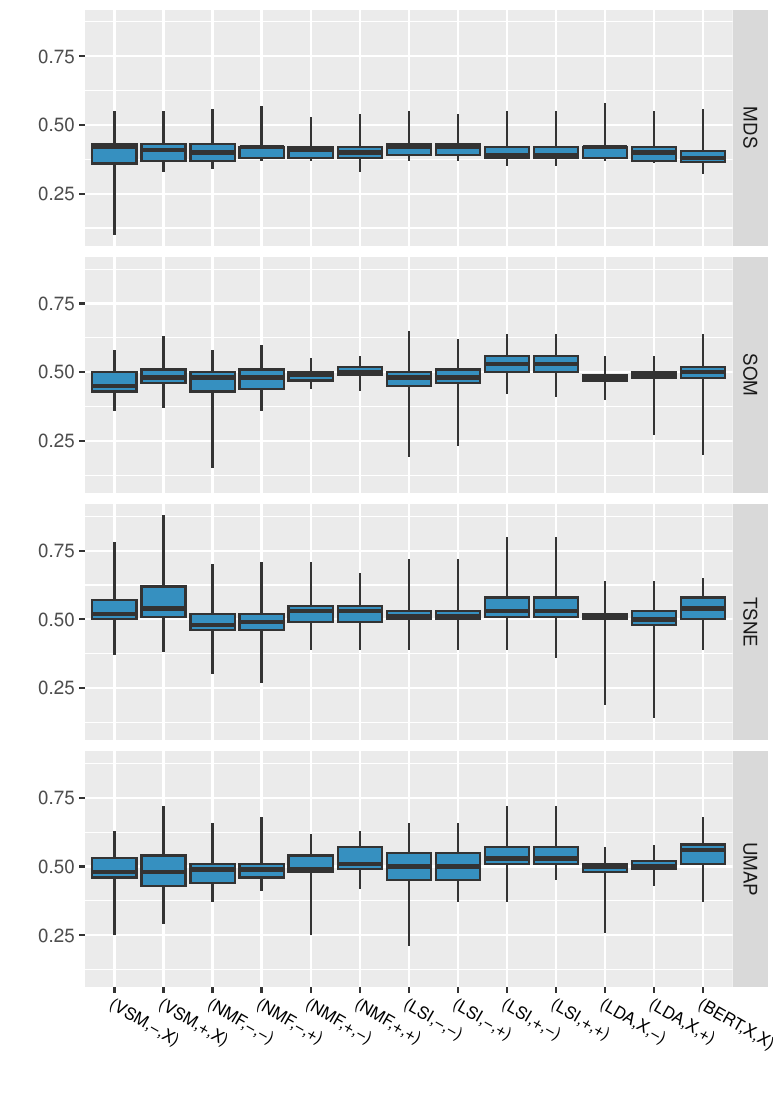}
    \vspace{-1.5\baselineskip}
    \caption{Boxplots showing the summary statistic of the quality metric $\beta$.}
    \label{fig:Boxplot_Perception}
\end{figure}

In addition to the optimal values, it is particularly relevant how sensitive the DRs are to their hyperparameters.
\Cref{fig:Boxplot_Accuracy} and \Cref{fig:Boxplot_Perception} show the five-number summaries for the quality metrics $\alpha$ and $\beta$ merged over the datasets.
The TMs are specified as a triple analogous to the description from above but with no indication of the DR.
\Cref{fig:Boxplot_Accuracy} shows that for all layout algorithms that rely on MDS, $\alpha$ takes on values within a small range. 
\Cref{fig:Boxplot_Perception} shows the same pattern with respect to $\beta$.
Even though a small range is desirable, MDS performs worse than the other DRs.
For layouts derived from a SOM, $\alpha$ varies within a large range.
The first and third quartiles are usually centered in the middle of the entire value range.
Therefore, it is difficult to achieve good results when using a SOM.
The range of values for $\beta$ is more restricted in most cases.
However, in cases of considerable variation, especially (LSI,-,-) and (LSI,-,+), the unfavorable location of the first and third quartiles can again be observed.
For t-SNE, $\alpha$ also varies within a wide range.
For all TMs, we observe that values above the first quartile are close to the optimum.
From this, we conclude that with a high probability, any chosen hyperparameter setting achieves good results.
Concerning $\beta$, the situation is different.
Values below the third quartile are usually close to the minimum.
Although the values for the first and third quartiles are similar to those of the SOMs, t-SNE shows more upside potential with respect to $\beta$.
For UMAP, we observe similar patterns as for t-SNE with respect to $\alpha$, albeit less pronounced and with apparent exceptions, e.g., (VSM,+,X).
Regarding $\beta$, the location of the first and third quartiles resembles those of t-SNE. 
Concerning $\beta$, UMAP, and t-SNE perform similarly.

\subsection{Performance of Default Hyperparameters}

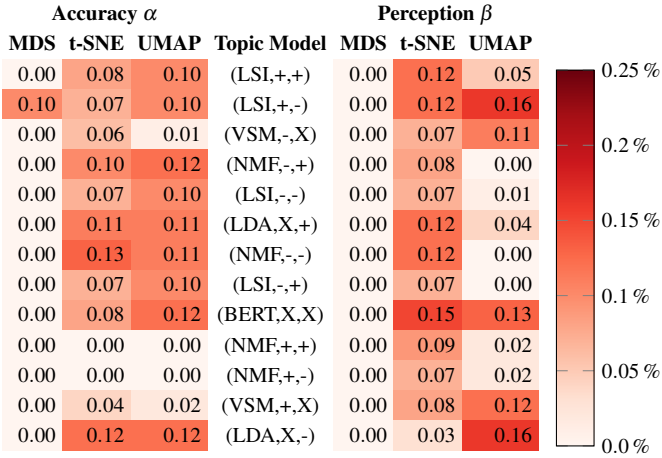
\begin{figure}[t]
    \centering
    %\vspace{-5pt}
    \pgfplotstableread[col sep=comma,skip first n=1,text indicator="]{data/Results_DefaultValues_Accuraries_Perception.csv}\loadeddata%
    \setlength{\tabcolsep}{2.5pt}%
    \renewcommand{\arraystretch}{1.2}%
    \pgfplotstabletypeset[%
        font=\footnotesize,
        col sep=&,
        row sep=\\,
        columns={
            MDS-a,
            TSNE-a,
            UMAP-a,
            TM,
            MDS-b,
            TSNE-b,
            UMAP-b,
            ColorBar
        },
        columns/TM/.style={
            string type,
            column type={c},
            column name={ \textbf{Topic Model} },
        },
        columns/MDS-a/.style={
            colorbarstyle,
            precision=2,
            column name={ \textbf{MDS} },
            color cells={min=0,max=0.25},
            /pgfplots/colormap name=Reds,
        },
        columns/MDS-b/.style={
            colorbarstyle,
            precision=2,
            column name={ \textbf{MDS} },
            color cells={min=0,max=0.25},
            /pgfplots/colormap name=Reds,
        },
        columns/TSNE-a/.style={
            colorbarstyle,
            precision=2,
            column name={ \textbf{t-SNE} },
            color cells={min=0,max=0.25},
            /pgfplots/colormap name=Reds,
        },
        columns/TSNE-b/.style={
            colorbarstyle,
            precision=2,
            column name={ \textbf{t-SNE} },
            color cells={min=0,max=0.25},
            /pgfplots/colormap name=Reds,
        },
        columns/UMAP-a/.style={
            colorbarstyle,
            precision=2,
            column name={ \textbf{UMAP} },
            color cells={min=0,max=0.25},
            /pgfplots/colormap name=Reds,
        },
        columns/UMAP-b/.style={
            colorbarstyle,
            precision=2,
            column name={ \textbf{UMAP} },
            color cells={min=0,max=0.25},
            /pgfplots/colormap name=Reds,
        },
        multicolumn names=c,
        every head row/.style={
            before row={
                \multicolumn{3}{c}{\textbf{Accuracy} $\alpha$} &
                \multicolumn{1}{c}{} &
                \multicolumn{3}{c}{\textbf{Perception} $\beta$} 
                \\
            },
        },
        create on use/ColorBar/.style={create col/set={}},
        columns/ColorBar/.style={
            column name={},
            assign cell content/.code={%
            \ifnum\pgfplotstablerow=0%
                \pgfkeyssetvalue{/pgfplots/table/@cell content}%
                {\multirow{\pgfplotstablerows}{*}{
                    \hspace*{0cm}
                    \begin{tikzpicture}[inner sep=0pt,outer sep=0pt]
                        \begin{axis}[
                            hide axis,
                            scale only axis,
                            height=0cm,
                            width=0cm,
                            colormap/Reds,
                            colorbar horizontal,
                            point meta min=0,
                            point meta max=0.25,
                            colorbar style={
                                width=5.0cm,
                                rotate=90,
                                xtick style={draw=gray},
                                xtick={0,0.05,0.1,0.15,0.2,0.25},
                                xticklabels={$0.0$\,\%,$0.05$\,\%,$0.1$\,\%,$0.15$\,\%,$0.2$\,\%,$0.25$\,\%},
                                xticklabel style={
                                    font=\footnotesize,
                                    xshift=3pt,
                                    yshift=-0.2ex,
                                    anchor=west,
                                },
                                typeset ticklabels with strut,
                                at={(0.0,0.0)},anchor=south,
                            }]
                            \addplot [draw=none] coordinates {(0,0) (1,1)};
                        \end{axis}
                    \end{tikzpicture}
                }}%
            \else
                \pgfkeyssetvalue{/pgfplots/table/@cell content}{}%
            \fi%
            },
        }
    ]{\loadeddata}
    \vspace{-0.5\baselineskip}
    \caption{Heatmap showing the percentage of resulting layouts that perform better than the default configuration for a given layout algorithm. (Left) Results according to the accuracy metric $\alpha$, (Right) Results according to the perception metric $\beta$.}
    \label{fig:Heatmap_DefaultValues}
\end{figure}

In typical application scenarios, the hyperparameters for a DR are left at their default values.
Therefore, we analyzed the quality scores of the respective layouts compared to the other scores within our dataset.
For each combination of TM, DR, and dataset, we determined the proportion of hyperparameter settings that yield better results for the metrics $\alpha$ and $\beta$ than the default settings.
The results, averaged over datasets, are summarized in \Cref{fig:Heatmap_DefaultValues} for MDS, t-SNE, and UMAP.
As the Sparse-SOM library has no default value for the number of neurons, we omitted this DR.
MDS only requires the specification of the number of iterations.
The implementation provided by Scikit-Learn specifies 300 as the default value.
Across all TMs, we observe values close to the optimum.
This indicates that MDS converges to a stable layout already after 300 iterations.
t-SNE is known for being sensitive to its hyperparameters.
Scikit-Learn specifies the default hyperparameters as $\texttt{perplexity} = 30$, $\texttt{n\_iter} = 1000$, and $\texttt{learning\_rate} = \max(m/48,50)$, where $m$ is the number of documents in our case.
Only \numprint[\%]{13} of the hyperparameter settings perform better than the default values with respect to $\alpha$.
With respect to $\beta$ only \numprint[\%]{15} achieve better results.
UMAP-Learn specifies the default values to $\texttt{n\_neighbors} = 15$ and $\texttt{min\_dist} = 0.1$.
In any case, maximal \numprint[\%]{13} of the layouts perform better than the default settings with respect to $\alpha$.
With respect to $\beta$, maximal \numprint[\%]{16} of the hyperparameter settings achieve better results.
Overall, the default values achieve good results.

\section{Discussion}
\label{section: discussion}

\revisedcontent{From the results of our evaluation, we derive user guidelines for the effective combination of TMs and DRs.}
However, our benchmark, and the guidelines strictly derived from it, are subject to threats to validity.

\subsection{Guidelines}
One of the main goals for two-dimensional layouts for text corpora is the preservation of structures within the high-dimensional representation, and the separability of clusters in the low-dimensional representation.
We captured both the accuracy metric $\alpha$ and the perception metric $\beta$, respectively.
\revisedcontent{The tf-idf weighting is often applied on the DTM as an additional preprocessing step.
Our first experiment indicates, on a formal statistical justification, that the tf-idf weighting tends to improve both the accuracy and the perception.}
\begin{enumerate}
    \item[\textbf{G1}] \revisedcontent{When applying the LSI, NMF, or no TM, the DTM should be weighted according to the tf-idf scheme.}
\end{enumerate}
\revisedcontent{Analogously our binary tests revealed that applying \Cref{eq: convex combination} improves the results.}
\begin{enumerate}
    \item[\textbf{G2}] \revisedcontent{When applying the LSI, NMF, or LDA, the document positions should be aggregated according to \Cref{eq: convex combination}.}
\end{enumerate}
\revisedcontent{For each dataset, we trained exactly one version for each TM.
Even without adjusting the hyperparameters of the TM, e.g., the number of topics, our second experiment showed that better results could be achieved when using a TM rather than solely relying on the VSM concerning $\alpha$.
Concerning $\beta$, a consecutive TM did not show improvements.
However, as we did not investigate the full capabilities of the TM, we can not conclude whether TM can improve the results concerning $\beta$.
Furthermore, the tf-idf weighting showed improvements concerning $\alpha$, and usually the tf-idf weighting results in better interpretable topics. We deduce:}
\begin{enumerate}
    \item[\textbf{G3}] \revisedcontent{A interpretable TM will probably improve the quality of a layout with respect to $\alpha$.}
\end{enumerate}
\revisedcontent{For both $\alpha$ and $\beta$ best results were achieved using t-SNE. 
Furthermore, our third experiment revealed that with respect to $\alpha$ most hyperparameter settings result in layouts near the optimum. The default values result in high-quality layouts.}
\begin{enumerate}
    \item[\textbf{G4}] \revisedcontent{We recommend the use of t-SNE as DR with its default values.}
\end{enumerate}
\revisedcontent{In particular, the use of t-SNE is also recommended by Nonato and Aupetit for the analytics task \enquote{Explore Items in Base Layout}~\cite{Nonato2019MDP} and previous benchmarks~\cite{Espadoto2021Toward}.}

\subsection{Threats to Validity}
Our design of the benchmark, its execution, the analysis of the results, and the derived guidelines are subject to internal and external threats to validity.
We identify two kinds of internal threats to validity.
The first concerns errors that may have occurred in evaluating the benchmark (Instrumentation).
For example, our results may be subject to human errors in software development.
For one, the risk is mitigated by using actively maintained open-source libraries for all algorithmic aspects that are often used in the ML community.
Further, the source code was reviewed by at least one additional co-author.
Last, we publish our entire implementation to allow for future improvements.
The second internal threat concerns errors that may be caused by the adjustments made during the execution of the benchmark (Attrition).
The layouts could be computed in $\approx$\numprint[\%]{94.7} of targeted cases.
Those cases can be attributed to exceeding memory consumption.
Even using the entire RAM of a node ($\approx$\numprint[GiB]{400}) has not prevented an out-of-memory event.

As the main external threat to validity, we identify the sampling bias.
Our benchmark is subject to sampling bias with respect to our selection of datasets, TMs, DRs, hyperparameters, their value range, and quality metrics.
In particular, it is unclear how our proposed guidelines are generalizable to other datasets according to the \emph{no free lunch theorem}~\cite{adam2019no}.
Furthermore, we trained one TM for each dataset with fixed hyperparameters.
Even though we manually viewed the resulting topics and followed best practices, it is unclear how the choices of hyperparameters for the TM, e.g., the number of topics, affect the results.
\revisedcontent{Furthermore, many of the layout algorithms sampled are not deterministic.
It is unclear to what extent this affects the generated layouts.
To mitigate this, we plan to evaluate multiple runs for a fixed hyperparameter configuration and evaluate average values with confidence intervals.}
Our choice of quality metrics was heavily influenced by previous benchmarking studies.
However, we did not consider the \emph{Normalized Stress} because, in many cases, it exceeds the range from 0 to 1, denoted as the value range in~\cite{Espadoto2021Toward,vernier2021guided}.
To address the sampling bias, the benchmark is designed to be extensible with respect to $\mathcal{D}$, $\mathcal{L}$, and $\mathcal{Q}$.
\revisedcontent{Furthermore, we see great potential in quantifying the quality of our benchmark, e.g., by measuring the \textit{Data Quality Index} proposed by Mishra et al.~\cite{mishra2020dqi}.}

\section{Conclusions \& Future Work}
\label{section: conclusions}

Many visualizations for text corpora rely on a two-dimensional spatialization derived from combining a TM and a subsequent DR.
Even though the choice of the TM, DR, and their respective hyperparameters significantly impacts the resulting layout, it is unknown how to obtain a two-dimensional layout reflecting both the structure within the corpus and the cluster separation between categories.
We proposed a benchmark $\mathcal{B} =(\mathcal{D}, \mathcal{L}, \mathcal{Q})$ consisting of a set of text corpora $\mathcal{D}$, a set of layout algorithms $\mathcal{L}$ that are combinations of TMs and DRs, and a set of quality metrics $\mathcal{Q}$.
We  published our benchmark, which is also designed to be extensible for further experiments.
By analyzing the correlation between the quality metrics, we defined an accuracy metric $\alpha$,  capturing  preservation of high-dimensional structures in the layout, and a perception metric $\beta$, capturing  separability between clusters.
By extensive analysis, we discussed the results after executing our benchmark and derived guidelines for the effective use of TMs and DRs for generating two-dimensional layouts for text corpora.
We recommend the use of LSI or the VSM, depending on whether the aggregated accuracy metric $\alpha$ or the aggregated perception metric $\beta$ is to be optimized.
The results can further be improved by applying the tf-idf weighting scheme.
In our experiments, t-SNE has shown the overall best performance.
In any case, the layout originating from the default hyperparameters is among the top \numprint[\%]{20}.
Unfortunately, none of the visualization approaches specified in \Cref{tab:vis techniques} used our recommended layout algorithm.
We hope practitioners and researchers consider our results and guidelines in their visualization design or extend the benchmark for a more  evaluation.

For future work, we see different promising directions.
We plan to expand our benchmark to address the major threats to validity.
Also, we see potential in evaluating more variants of a TM, i.e., by iterating over its hyperparameter, \revisedcontent{or new variants of the BERT model, e.g., specialized for the case of source code.}
Furthermore, the temporal stability of a layout is particularly interesting for streaming text corpora,  e.g., from social media.
Our  benchmark could be extended to incorporate time stability as proposed by Vernier et al.~\cite{vernier2021guided}.
In addition to the quality metrics, the layouts are saved as well.
\revisedcontent{We plan to analyze the relationship between DRs and their generated shapes in the two-dimensional representation.
For this, we measure metrics related to the shape, e.g., the popular scagnostics~\cite{wilkinson2005graph,wilkinson2006HighDimensional} or measures presented by Xia et al.~\cite{Xia2021VisualClustering}.
This results in a dataset where each point describes the shape of a single scatter plot. 
By clustering this dataset and labeling the categories according to their visual characteristics, one can determine which DR will likely result in a specific shape, as recently presented by Machado et al.~\cite{machado2023-sharp}, and whether it is beneficial for a particular text visualization task.}
It would also be interesting to do a coding of resulting scatter plot patterns by means of an open coding study \cite{DBLP:conf/chi/PandeyKFBB16}.
\revisedcontent{For a large corpus, computational costs are  relevant.
We consider evaluating the runtime of different layout algorithms.}
To improve computation times for layouts, neural networks can be trained on a set of precomputed layouts to approximate a given DR~\cite{espadoto2020deep}.
This, in particular, enables applications such as  interactive exploration of the hyperparameter space of DR~\cite{appleby2022hypernp} or comparative analysis~\cite{Fujiwara2022InteractiveDR}.
We plan to analyze to what extent our dataset is suitable for training a neural network that predicts layouts based on TMs for text corpora.

\section*{Acknowledgements}
We thank the anonymous reviewers for their valuable feedback to improve this work.
This work was partially funded by the Federal Ministry of Education and Research, Germany through grant 01IS22062 (\enquote{AI research group FFS-AI}), and grant 01IS20088B (\enquote{KnowhowAnalyzer}).
Furthermore, this work is part of project 16KN086467 (\enquote{DecodingFood}) funded by the Federal Ministry for Economic Affairs and Climate Action of Germany.
The work of Tobias Schreck was partially funded by the Austrian Research Promotion Agency (FFG) within the framework of the flagship project ICT of the Future PRESENT, grant FO999899544.

\bibliographystyle{abbrv-doi-hyperref-narrow}

\bibliography{main}

\appendix



\section*{Supplementary material}

# Large-Scale Evaluation of Topic Models and Dimensionality Reduction Methods for 2D Text Spatialization
## Supplemental Material

Daniel Atzberger, Tim Cech, Matthias Trapp, Rico Richter,
Willy Scheibel, Jürgen Döllner, and Tobias Schreck

## A  Topics

Table 1 – Table 29 show the top ten words for selected topics of all considered Topic Models.

## B  Example Layouts

Figure 1 shows the layouts generated from (LSI,+,TSNE,+) and (VSM,+,TSNE,X) for all datasets respectively using the default parameters of t-SNE. The color represents the class label.

Table 1: Top ten words for ten selected topics for the LSI model for the 20 Newsgroups dataset.

| 0 | 1 | 2 | 3 | 4 | 5 | 6 | 7 | 8 | 9 |
|---|---|---|---|---|---|---|---|---|---|
| would | window | key | key | drive | god | card | drive | car | space |
| use | file | game | god | window | gun | driver | gun | window | file |
| get | god | chip | game | file | car | drive | window | bike | henry |
| people | card | team | team | ide | card | file | file | god | gun |
| one | drive | clipper | chip | card | key | gun | ide | key | sale |
| say | driver | encryption | clipper | controller | game | video | disk | driver | zoo |
| window | people | player | encryption | card | drive | disk | sale | gun | window |
| know | disk | win | player | disk | ide | window | mail | space | car |
| go | video | play | window | program | bike | mouse | monitor | mouse | god |
| think | color | escrow | escrow | win | chip | diamond | god | ride | mail |

Table 2: Top ten words for the eight topics for the LSI model for the Emails dataset.

| 0 | 1 | 2 | 3 | 4 | 5 | 6 | 7 |
|---|---|---|---|---|---|---|---|
| key | key | bit | key | stratus | phone | access | gun |
| chip | chip | government | stratus | transfer | tap | net | genocide |
| encryption | clipper | encryption | bit | key | public | pat | turk |
| government | encryption | space | net | rocket | clipper | key | chip |
| clipper | escrow | key | access | gun | line | stratus | clipper |
| use | bit | law | encryption | message | encryption | encryption | soviet |
| would | gun | stratus | gun | bit | serial | chip | stratus |
| people | algorithm | clipper | clipper | distribution | warrant | escrow | firearm |
| phone | phone | gun | technology | carl | file | reference | bony |
| system | serial | privacy | host | encrypt | security | space | law |

Table 3: Top ten words for ten selected topics for the LSI model for the GitHub projects dataset.

| 0 | 1 | 2 | 3 | 4 | 5 | 6 | 7 | 8 | 9 |
|---|---|---|---|---|---|---|---|---|---|
| texture | texture | tensor | webpack | price | mut | mut | wallet | webpack | react |
| prototype | vec | torch | react | market | webpack | webpack | blockchain | react | webpack |
| vec | shader | train | chart | candle | pub | pub | transaction | err | err |
| webpack | vertex | np | vue | trade | err | chart | mut | syscall | jest |
| vertex | mesh | webpack | pub | exchange | wallet | err | boost | sqlite | sqlite |
| mesh | err | shape | mut | buy | react | plot | err | plot | vue |
| tensor | gl | tf | vec | binance | crate | crate | chart | jest | syscall |
| model | geometry | react | texture | mut | func | axis | std | std | eslint |
| shader | material | err | wallet | pub | chart | fn | pub | prop | sql |
| std | tensor | model | err | symbol | vec | react | block | vue | chart |

Table 4: Top ten words for ten selected topics for the LSI model for the Reuters dataset.

| 0 | 1 | 2 | 3 | 4 | 5 | 6 | 7 | 8 | 9 |
|---|---|---|---|---|---|---|---|---|---|
| loss | loss | net | div | billion | profit | share | bank | split | split |
| net | net | div | billion | share | loss | billion | rate | dividend | gain |
| profit | billion | record | prior | profit | billion | split | oil | quarter | quarter |
| nine | profit | loss | record | stock | tax | stock | sale | rise | oil |
| year | div | prior | pay | div | bank | rate | prime | offer | stock |
| sale | record | pay | net | offer | loan | bank | split | year | exclude |
| note | prior | dividend | bank | net | nil | dividend | money | corp | crude |
| billion | nine | set | quarterly | rise | net | common | price | group | sale |
| include | pay | march | dividend | say | share | oil | year | stake | include |
| gain | sale | quarterly | say | common | three | rise | billion | div | tax |

Table 5: Top ten words for ten selected topics for the LSI model for the Seven Categories dataset.

| 0 | 1 | 2 | 3 | 4 | 5 | 6 | 7 | 8 | 9 |
|---|---|---|---|---|---|---|---|---|---|
| instruction | share | share | war | share | cache | cell | debenture | charge | fraction |
| register | instruction | debenture | charge | debenture | instruction | cache | cash | motion | decimal |
| memory | register | allotment | soviet | cash | miss | debenture | sale | wave | number |
| bit | debenture | instruction | magnetic | purchase | block | plant | account | magnetic | bit |
| address | memory | cash | united | account | register | cash | cell | particle | lesson |
| cache | cash | charge | party | sale | memory | charge | purchase | field | exponent |
| branch | bit | application | government | allotment | page | water | plant | electric | denominator |
| charge | address | money | president | balance | branch | organism | redemption | current | instruction |
| cycle | allotment | capital | field | redemption | charge | miss | error | velocity | multiply |
| clock | charge | force | electric | book | hierarchy | magnetic | issue | cell | review |

Table 6: Top ten words for ten selected topics for the LSI model with tfidf weighting for the 20 Newsgroups dataset.

| 0 | 1 | 2 | 3 | 4 | 5 | 6 | 7 | 8 | 9 |
|---|---|---|---|---|---|---|---|---|---|
| would | window | key | key | drive | god | card | drive | car | space |
| use | file | game | god | window | gun | driver | gun | window | file |
| get | god | chip | game | file | car | drive | window | bike | henry |
| people | card | team | team | ide | card | file | file | god | gun |
| one | drive | clipper | chip | card | key | gun | ide | key | sale |
| say | driver | encryption | clipper | controller | game | video | disk | driver | zoo |
| window | people | player | encryption | card | drive | disk | sale | gun | window |
| know | disk | win | player | disk | ide | window | mail | space | car |
| go | video | play | window | program | bike | mouse | monitor | mouse | god |
| think | color | escrow | escrow | win | chip | diamond | god | ride | mail |

Table 7: Top ten words for the eight topics for the LSI model with tfidf weighting for the Emails dataset.

| 0 | 1 | 2 | 3 | 4 | 5 | 6 | 7 |
|---|---|---|---|---|---|---|---|
| key | key | bit | key | stratus | phone | access | gun |
| chip | chip | government | stratus | transfer | tap | net | genocide |
| encryption | clipper | encryption | bit | key | public | pat | turk |
| government | encryption | space | net | rocket | clipper | key | chip |
| clipper | escrow | key | access | gun | line | stratus | clipper |
| use | bit | law | encryption | message | encryption | encryption | soviet |
| would | gun | stratus | gun | bit | serial | chip | stratus |
| people | algorithm | clipper | clipper | distribution | warrant | escrow | firearm |
| phone | phone | gun | technology | carl | file | reference | bony |
| system | serial | privacy | host | encrypt | security | space | law |

Table 8: Top ten words for ten selected topics for the LSI model with tfidf weighting for the GitHub projects dataset.

| 0 | 1 | 2 | 3 | 4 | 5 | 6 | 7 | 8 | 9 |
|---|---|---|---|---|---|---|---|---|---|
| texture | texture | tensor | webpack | price | mut | mut | wallet | webpack | react |
| prototype | vec | torch | react | market | webpack | webpack | blockchain | react | webpack |
| vec | shader | train | chart | candle | pub | pub | transaction | err | err |
| webpack | vertex | np | vue | trade | err | chart | mut | syscall | jest |
| vertex | mesh | webpack | pub | exchange | wallet | err | boost | sqlite | sqlite |
| mesh | err | shape | mut | buy | react | plot | err | plot | vue |
| tensor | gl | tf | vec | binance | crate | crate | chart | jest | syscall |
| model | geometry | react | texture | mut | func | axis | std | std | eslint |
| shader | material | err | wallet | pub | chart | fn | pub | prop | sql |
| std | tensor | model | err | symbol | vec | react | block | vue | chart |

Table 9: Top ten words for ten selected topics for the LSI model with tfidf weighting for the Reuters dataset.

| 0 | 1 | 2 | 3 | 4 | 5 | 6 | 7 | 8 | 9 |
|---|---|---|---|---|---|---|---|---|---|
| loss | loss | net | div | billion | profit | share | bank | split | split |
| net | net | div | billion | share | loss | billion | rate | dividend | gain |
| profit | billion | record | prior | profit | billion | split | oil | quarter | quarter |
| nine | profit | loss | record | stock | tax | stock | sale | rise | oil |
| year | div | prior | pay | div | bank | rate | prime | offer | stock |
| sale | record | pay | net | offer | loan | bank | split | year | exclude |
| note | prior | dividend | bank | net | nil | dividend | money | corp | crude |
| billion | nine | set | quarterly | rise | net | common | price | group | sale |
| include | pay | march | dividend | say | share | oil | year | stake | include |
| gain | sale | quarterly | say | common | three | rise | billion | div | tax |

Table 10: Top ten words for ten selected topics for the LSI model with tfidf weighting for the Seven Categories dataset.

| 0 | 1 | 2 | 3 | 4 | 5 | 6 | 7 | 8 | 9 |
|---|---|---|---|---|---|---|---|---|---|
| instruction | share | share | war | share | cache | cell | debenture | charge | fraction |
| register | instruction | debenture | charge | debenture | instruction | cache | cash | motion | decimal |
| memory | register | allotment | soviet | cash | miss | debenture | sale | wave | number |
| bit | debenture | instruction | magnetic | purchase | block | plant | account | magnetic | bit |
| address | memory | cash | united | account | register | cash | cell | particle | lesson |
| cache | cash | charge | party | sale | memory | charge | purchase | field | exponent |
| branch | bit | application | government | allotment | page | water | plant | electric | denominator |
| charge | address | money | president | balance | branch | organism | redemption | current | instruction |
| cycle | allotment | capital | field | redemption | charge | miss | error | velocity | multiply |
| clock | charge | force | electric | book | hierarchy | magnetic | issue | cell | review |

Table 11: Top ten words for ten selected topics for the LDA model for the 20 Newsgroups dataset.

| 0 | 1 | 2 | 3 | 4 | 5 | 6 | 7 | 8 | 9 |
|---|---|---|---|---|---|---|---|---|---|
| gun | say | would | subject | use | use | drive | go | game | god |
| space | people | write | line | line | line | organization | say | team | church |
| year | one | think | organization | window | subject | subject | one | year | henry |
| weapon | know | subject | system | subject | organization | line | would | player | believe |
| firearm | would | article | post | organization | get | get | get | good | subject |
| orbit | time | like | book | file | need | university | people | subject | say |
| control | see | organization | use | card | one | post | know | organization | line |
| rate | think | line | mail | problem | also | would | well | line | organization |
| use | come | make | university | bit | look | write | make | write | one |
| new | write | say | computer | get | write | article | think | get | write |

Table 12: Top ten words for the eight topics for the LDA model for the Emails dataset.

| 0 | 1 | 2 | 3 | 4 | 5 | 6 | 7 |
|---|---|---|---|---|---|---|---|
| write | write | people | president | would | homosexual | would | would |
| gay | article | write | government | one | write | use | write |
| say | post | right | say | make | one | one | one |
| think | get | say | new | go | use | government | year |
| one | say | gun | people | write | article | law | article |
| go | would | article | go | people | bit | make | make |
| use | go | would | use | know | health | key | soviet |
| see | host | think | human | think | get | write | use |
| article | one | one | would | article | know | people | apartment |
| people | fire | go | write | say | would | say | million |

Table 13: Top ten words for ten selected topics for the LDA model for the GitHub projects dataset.

| 0 | 1 | 2 | 3 | 4 | 5 | 6 | 7 | 8 | 9 |
|---|---|---|---|---|---|---|---|---|---|
| get | get | view | type | type | type | builder | wx | resource | sph |
| webpack | type | value | boost | webpack | get | info | model | name | gl |
| value | string | pyx | get | get | value | get | i | get | get |
| type | value | get | err | node | model | protobuf | d | value | value |
| node | size | object | value | order | size | com | string | xamarin | type |
| set | node | set | name | i | data | outer | get | set | set |
| name | set | type | string | module | name | grasscutter | std | type | index |
| key | name | insert | i | io | gl | net | mm | ly | size |
| ag | std | d | d | value | input | google | item | droid | name |
| i | d | name | set | d | set | emu | event | wd | string |

Table 14: Top ten words for ten selected topics for the LDA model for the Reuters dataset.

| 0 | 1 | 2 | 3 | 4 | 5 | 6 | 7 | 8 | 9 |
|---|---|---|---|---|---|---|---|---|---|
| oil | stock | billion | say | say | rate | say | say | price | price |
| energy | say | profit | rate | company | debenture | talk | would | raise | official |
| say | cocoa | net | currency | year | standard | fire | official | oil | japan |
| pipeline | buffer | year | market | quarter | effective | ship | oil | crude | output |
| barrel | delegate | loss | dollar | earning | say | preliminary | meeting | barrel | chip |
| crude | buy | sale | bank | share | split | company | make | soviet | say |
| day | manager | company | west | first | discount | south | plan | increase | reed |
| spot | rule | note | exchange | expect | plan | cause | take | say | cut |
| ford | consumer | nine | monetary | report | treasury | bankruptcy | import | post | market |
| york | purchase | operating | german | result | balance | report | country | west | fall |

Table 15: Top ten words for ten selected topics for the LDA model for the Seven Categories dataset.

| 0 | 1 | 2 | 3 | 4 | 5 | 6 | 7 | 8 | 9 |
|---|---|---|---|---|---|---|---|---|---|
| number | page | mass | city | cache | address | bit | execution | clock | instruction |
| point | hazard | energy | state | memory | force | reg | program | instruction | memory |
| magnetic | memory | nucleus | war | miss | block | register | branch | use | time |
| field | performance | atom | new | datum | particle | one | file | cycle | pipeline |
| two | use | two | also | use | body | use | designer | one | bit |
| use | time | error | two | write | system | time | address | figure | computer |
| force | program | electron | country | time | register | figure | system | memory | motion |
| time | code | decay | form | address | mass | two | also | state | use |
| vector | system | neutron | wave | register | data | also | force | exception | load |
| direction | force | page | south | water | procedure | call | state | two | two |

Table 16: Top ten words for ten selected topics for the NMF model for the 20 Newsgroups dataset.

| | 0 | 1 | 2 | 3 | 4 | 5 | 6 | 7 | 8 | 9 |
|---|---|---|---|---|---|---|---|---|---|---|
| | use | ax | god | use | file | would | health | year | ax | file |
| | argument | di | write | say | gun | one | would | good | di | image |
| | one | ey | subject | one | firearm | people | center | line | um | mail |
| | drive | biz | people | time | control | say | medical | use | ey | use |
| | make | ex | organization | people | bill | know | use | output | mu | send |
| | fallacy | mu | article | year | handgun | make | os | well | ah | graphic |
| | would | ne | father | health | law | think | cancer | write | mi | gun |
| | post | mi | say | get | united | write | mu | use | ex | format |
| | see | om | son | work | weapon | well | tobacco | program | pu | system |
| | problem | ah | make | go | crime | use | new | file | biz | ray |

Table 17: Top ten words for the eight topics for the NMF model for the Emails dataset.

| | 0 | 1 | 2 | 3 | 4 | 5 | 6 | 7 |
|---|---|---|---|---|---|---|---|---|
| | one | may | say | anonymous | health | use | privacy | key |
| | say | address | go | use | year | law | file | bit |
| | go | anonymous | president | service | disease | government | one | one |
| | people | user | know | posting | child | encryption | information | planet |
| | would | people | think | post | number | chip | network | first |
| | get | anonymity | work | user | report | technology | electronic | block |
| | come | system | well | anon | medical | new | security | earth |
| | see | right | take | server | state | clipper | mail | chip |
| | like | identity | make | message | patient | would | public | message |
| | could | many | key | version | study | key | policy | system |

Table 18: Top ten words for ten selected topics for the NMF model for the GitHub projects dataset.

| | 0 | 1 | 2 | 3 | 4 | 5 | 6 | 7 | 8 | 9 |
|---|---|---|---|---|---|---|---|---|---|---|
| | type | ly | type | type | type | get | flow | flow | flow | flow |
| | boost | get | flow | boost | flow | ly | direction | type | type | direction |
| | get | qp | qp | get | value | boost | type | ptr | name | type |
| | value | type | name | value | get | qp | uint | direction | direction | ptr |
| | size | vl | ptr | d | name | vl | name | ptr | boost | uint |
| | result | wd | value | pp | ptr | wd | type | get | ptr | name |
| | pp | d | d | op | i | direction | value | uint | get | get |
| | string | value | uint | result | uint | set | count | d | uint | count |
| | d | i | size | string | pp | ul | summary | i | pp | summary |
| | set | err | direction | set | value | d | vtbl | ly | value | vtbl |

Table 19: Top ten words for ten selected topics for the NMF model for the Reuters dataset.

| | 0 | 1 | 2 | 3 | 4 | 5 | 6 | 7 | 8 | 9 |
|---|---|---|---|---|---|---|---|---|---|---|
| | company | source | profit | currency | march | say | year | price | rise | barrel |
| | say | oil | loss | exchange | week | billion | last | new | fall | crude |
| | corp | official | net | market | rise | sale | one | pact | year | west |
| | unit | share | share | say | fed | end | price | rubber | month | raise |
| | business | crude | note | foreign | compare | gas | rubber | world | adjust | grade |
| | sell | tax | nine | mark | quarter | expect | growth | cent | index | posting |
| | expect | last | include | rate | bill | total | market | reference | output | oil |
| | operation | could | company | system | gain | early | economic | conference | production | contract |
| | year | export | sale | bank | first | rise | expect | may | seasonally | petroleum |
| | product | plan | gain | west | drop | low | economy | month | say | petroleum |

Table 20: Top ten words for ten selected topics for the NMF model for the Seven Categories dataset.

| | 0 | 1 | 2 | 3 | 4 | 5 | 6 | 7 | 8 | 9 |
|---|---|---|---|---|---|---|---|---|---|---|
| | author | war | force | share | force | charge | water | energy | share | vector |
| | book | cash | motion | call | use | wave | surface | particle | light | cell |
| | write | soviet | two | chromosome | field | body | point | per | united | use |
| | movement | also | body | capital | time | electric | air | time | new | also |
| | include | would | state | figure | power | point | pressure | mass | world | system |
| | new | united | change | two | memory | time | plant | share | call | human |
| | many | government | law | issue | war | two | ocean | point | war | organism |
| | soviet | union | velocity | become | friction | direction | call | work | account | time |
| | national | new | acceleration | year | would | give | due | potential | use | two |
| | country | become | fig | population | china | surface | movement | surface | also | animal |

Table 21: Top ten words for ten selected topics for the NMF model with tfidf weighting for the 20 Newsgroups dataset.

| | 0 | 1 | 2 | 3 | 4 | 5 | 6 | 7 | 8 | 9 |
|---|---|---|---|---|---|---|---|---|---|---|
| | sun | food | chip | god | people | space | file | drive | thank | team |
| | de | arbor | clipper | believe | kill | government | window | disk | please | player |
| | objective | ann | encryption | faith | turk | bony | win | floppy | file | game |
| | tu | thing | escrow | atheist | say | jake | run | hard | mail | hockey |
| | font | disease | boston | hell | right | president | program | boot | advance | play |
| | graphic | name | key | say | fire | center | directory | problem | anyone | season |
| | value | mark | algorithm | atheism | attack | launch | manager | controller | duke | league |
| | display | mi | machine | existence | turkey | book | server | monitor | hi | win |
| | motif | question | privacy | belief | country | shuttle | buffalo | switch | know | ca |
| | program | level | government | exist | war | moon | version | se | state | ranger |

Table 22: Top ten words for the eight topics for the NMF model with tfidf weighting for the Emails dataset.

| | 0 | 1 | 2 | 3 | 4 | 5 | 6 | 7 |
|---|---|---|---|---|---|---|---|---|
| | clipper | chip | phone | get | pat | space | government | key |
| | encryption | device | trust | drug | bit | stratus | algorithm | bit |
| | phone | encryption | government | go | access | university | encryption | public |
| | key | algorithm | security | want | net | mail | patent | session |
| | chip | use | president | fire | run | post | secret | encrypt |
| | system | clipper | privacy | good | give | news | people | escrow |
| | escrow | voice | ensure | tell | thing | please | right | number |
| | use | technology | care | system | two | world | public | serial |
| | right | datum | omission | would | think | thank | law | chip |
| | criminal | privacy | law | gun | go | year | say | block |

Table 23: Top ten words for ten selected topics for the NMF model with tfidf weighting for the GitHub projects dataset.

| | 0 | 1 | 2 | 3 | 4 | 5 | 6 | 7 | 8 | 9 |
|---|---|---|---|---|---|---|---|---|---|---|
| | candle | table | sql | vue | react | wallet | webpack | wallet | pyx | chart |
| | shardingsphere | db | summary | texture | node | mut | plot | gl | wallet | sqlite |
| | glew | column | sprite | webpack | model | std | torch | train | serie | torch |
| | prop | client | player | xonsh | style | pub | kwargs | axis | err | price |
| | price | i | span | tensor | state | ua | np | tx | vtk | tensor |
| | trade | func | candle | neo | axis | ssh | train | chart | glew | exchange |
| | xonsh | err | engine | vertex | font | prism | trino | plot | vulkan | std |
| | apache | license | node | blockchain | value | orbitdb | numref | block | pub | market |
| | webpacker | d | hljs | camera | event | onig | plt | sph | mut | mut |
| | trading | np | puma | vec | get | plt | shiv | layer | crate | license |

Table 24: Top ten words for ten selected topics for the NMF model with tfidf weighting for the Reuters dataset.

| | 0 | 1 | 2 | 3 | 4 | 5 | 6 | 7 | 8 | 9 |
|---|---|---|---|---|---|---|---|---|---|---|
| | give | sugar | sugar | currency | billion | acquisition | output | franklin | buy | mine |
| | corp | rubber | white | baker | net | acquire | production | corp | share | dome |
| | assistance | yr | rebate | dollar | rise | common | china | fund | common | offer |
| | band | per | trader | stability | mark | complete | chip | income | stake | bid |
| | class | oil | intervention | exchange | yen | tender | month | insure | exchange | south |
| | public | market | export | franklin | year | pacific | life | free | march | debt |
| | go | extraordinary | declare | west | surplus | stock | sugar | board | group | gold |
| | help | consolidated | tender | industrial | deposit | company | cotton | march | agree | miner |
| | late | cattle | cargo | finance | end | hotel | period | tax | security | copper |
| | shortage | gas | dividend | treasury | reserve | prime | lynch | bell | commission | mining |

Table 25: Top ten words for ten selected topics for the NMF model with tfidf weighting for the Seven Categories dataset.

| | 0 | 1 | 2 | 3 | 4 | 5 | 6 | 7 | 8 | 9 |
|---|---|---|---|---|---|---|---|---|---|---|
| | plate | cell | cash | water | share | united | debenture | charge | wave | specie |
| | motion | inequality | decimal | pressure | nucleus | state | issue | magnetic | light | water |
| | body | plant | share | share | nuclear | soviet | party | field | electron | wave |
| | particle | lesson | digit | charge | energy | pressure | premium | force | speed | plant |
| | system | review | number | segment | equity | war | company | energy | acceleration | energy |
| | capacitor | instruction | percent | surface | neutron | new | government | electric | velocity | light |
| | rigid | chromosome | pay | wire | capital | return | share | vector | cash | fraction |
| | force | stage | round | company | decay | economic | allotment | mass | motion | organism |
| | axis | figure | cheque | tissue | reactor | president | election | point | object | diversity |
| | rotation | transport | point | blood | application | world | reserve | surface | statement | ecosystem |

Table 26: Top ten words for ten selected topics for the BERT model for the 20 Newsgroups dataset.

| 0 | 1 | 2 | 3 | 4 | 5 | 6 | 7 | 8 | 9 |
|---|---|---|---|---|---|---|---|---|---|
| file | dog | food | patient | nuclear | space | president | jake | marriage | buffalo |
| drive | chase | eat | disease | cool | launch | job | bony | marry | ra |
| state | bike | diet | doctor | tower | orbit | secretary | slave | married | ye |
| computer | driveway | seizure | medical | plant | moon | package | true | ceremony | god |
| want | ride | sensitivity | pain | water | satellite | administration | opinion | divorce | paradise |
| right | springer | taste | cancer | steam | henry | russia | russia | wife | father |
| window | questor | superstition | treatment | uranium | lunar | senior | bush | husband | salvation |
| thing | dod | corn | infection | reactor | mission | senator | slavery | couple | child |
| problem | attack | restaurant | health | fossil | shuttle | summer | prison | eye | thou |
| host | boom | dyer | yeast | temperature | earth | press | aid | commitment | mormon |

Table 27: Top ten words for the eight topics for the BERT model for the Emails dataset.

| 0 | 1 | 2 | 3 | 4 | 5 | 6 | 7 |
|---|---|---|---|---|---|---|---|
| gun | encryption | tap | bit | secrecy | trial | apple | agency |
| state | entitle | betel | sequence | agora | thousand | trust | safeguard |
| come | effective | traffic | chunk | rain | microsecond | credibility | escrow |
| right | threaten | prefix | actual | den | engine | secret | expedient |
| party | unbreakable | blank | split | repository | skipjack | forever | congress |
| war | safety | encryption | console | door | hypothesize | bu | funded |
| host | illegal | accuse | competence | variant | arithmetic | assurance | overtly |
| want | privacy | decode | random | security | brute | unsound | federal |
| law | saying | unauthorized | partial | instant | steven | steal | fund |
| world | outright | technique | randomly | mnemonic | million | connect | assurance |

Table 28: Top ten words for ten selected topics for the BERT model for the Reuters dataset.

| 0 | 1 | 2 | 3 | 4 | 5 | 6 | 7 | 8 | 9 |
|---|---|---|---|---|---|---|---|---|---|
| bank | mar | money | canada | sale | net | sale | corp | sale | net |
| rate | park | assistance | dome | net | year | net | net | net | roto |
| sale | net | shortage | trade | diluted | renaissance | diluted | rev | mercantile | yr |
| corp | zone | market | bank | dilute | pembina | year | odeon | crain | lighting |
| quarter | faulty | band | shell | corp | oe | vanguard | year | potlatch | dunk |
| stock | favorite | fed | rise | gem | charming | suspension | basic | servo | analogic |
| new | favorably | bank | billion | sexton | tempo | note | wallpaper | electromagnetic | dorn |
| end | favorable | reserve | air | ply | vanguard | backlog | sidy | whirlpool | napa |
| note | favor | forecast | fall | seam | foremost | end | sidy | translation | rooter |
| sell | fault | today | royal | burr | kay | rivet | gas | fuller | morse |

Table 29: Top ten words for seven selected topics for the BERT model for the Seven Categories dataset.

| 0 | 1 | 2 | 3 | 4 | 5 | 6 |
|---|---|---|---|---|---|---|
| space | cash | cache | motion | wind | united | plant |
| population | debenture | bit | particle | tide | soviet | organism |
| mission | account | address | fig | air | war | cell |
| satellite | share | register | magnetic | ocean | president | animal |
| pregnancy | company | processor | electric | warm | party | specie |
| moon | allotment | memory | wave | condensation | government | tissue |
| craft | purchase | computer | electron | cloud | union | blood |
| launch | capital | page | mass | cyclone | country | biology |
| exploration | balance | datum | velocity | tropical | political | acid |
| orbit | pay | clock | direction | circulation | military | protein |

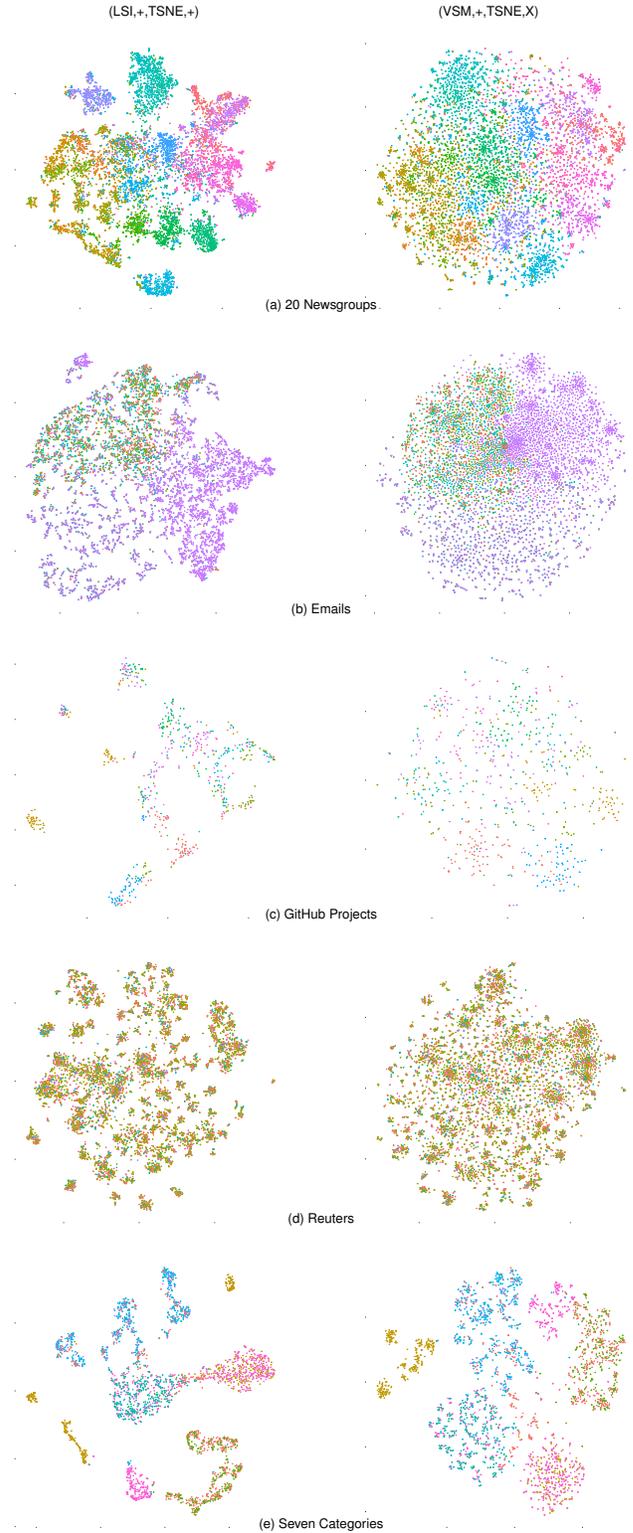

(a) 20 Newsgroups

(b) Emails

(c) GitHub Projects

(d) Reuters

(e) Seven Categories

Figure 1: Layout examples for LSI and VSM with applied t-SNE dimensionality reduction for the five datasets.



\end{document}